\documentclass[10pt,journal,compsoc]{IEEEtran}

\usepackage{amsmath}
\usepackage{amssymb}
\ifCLASSOPTIONcompsoc
  \usepackage[nocompress]{cite}
\else
  \usepackage{cite}
\fi
\usepackage{graphicx}
\usepackage{epsfig}
\usepackage{color}
\usepackage{algorithmic,algorithm}
\usepackage{url}
\usepackage{amsopn}
\usepackage{booktabs}

\def\bg#1{\mbox{\boldmath$#1$}} 

\DeclareMathOperator*{\argmax}{argmax}

\DeclareMathOperator*{\rank}{rank}
\DeclareMathOperator*{\trace}{tr}
\DeclareMathOperator*{\diag}{diag}
\DeclareMathOperator*{\rangesp}{range}

\DeclareMathOperator*{\spanvs}{span} 
\let\oldin\in

\def\in{\mathrel{
   \mathchoice{\raise.2ex\hbox{$\scriptstyle\oldin$}}
   {\raise.2ex\hbox{$\scriptstyle\oldin$}}
   {\raise.2ex\hbox{$\scriptscriptstyle\oldin$}}
   {\raise.2ex\hbox{$\scriptscriptstyle\oldin$}}}}

\newtheorem{theorem}{Theorem}[section]
\newtheorem{lemma}[theorem]{Lemma}

\begin{document}

\title{Accelerated kernel discriminant analysis}

\author{Nikolaos~Gkalelis,
        Vasileios~Mezaris,~\IEEEmembership{Senior~Member,~IEEE}
\IEEEcompsocitemizethanks{\IEEEcompsocthanksitem The authors are with Information Technologies
Institute, CERTH, 6th Km Charilaou-Thermi Road, P.O. BOX 60361,
Thermi 57001, Greece. \mbox{Tel.: ++ 30 2311257770}, \mbox{Fax: ++
30 2310474128}, E-mail: \{gkalelis, bmezaris\}@iti.gr \protect\\
}
\thanks{Manuscript received xxx; revised xxx.}}

\IEEEtitleabstractindextext{%
\begin{abstract}
In this paper, using a novel matrix factorization and simultaneous reduction to diagonal form approach 
(or in short simultaneous reduction approach),
Accelerated Kernel Discriminant Analysis (AKDA) and 
Accelerated Kernel Subclass Discriminant Analysis (AKSDA) are proposed.
Specifically, instead of performing the simultaneous reduction of the between- and within-class or subclass scatter matrices,
the nonzero eigenpairs (NZEP) of the so-called core matrix, which is of relatively small dimensionality, and
the Cholesky factorization of the kernel matrix are computed,
achieving more than one order of magnitude speed up over kernel discriminant analysis (KDA).
Moreover, consisting of a few elementary matrix operations and very stable numerical algorithms, AKDA and AKSDA
offer improved classification accuracy.
The experimental evaluation on various datasets confirms that
the proposed approaches provide state-of-the-art performance in terms of both training time and classification accuracy.
\end{abstract}

}

\maketitle

\IEEEdisplaynontitleabstractindextext

\IEEEpeerreviewmaketitle

\IEEEraisesectionheading{\section{Introduction} \label{S:Introduction}}

\IEEEPARstart{D}{imensionality} reduction (DR) using discriminant analysis (DA) \cite{Fukunaga90,YanPami07,Zhu06,You11,YuPami12,Gkalelis13NNLS,HePami16,SuPami16}
is a popular analytical framework with broad applications in
event and concept detection in video \cite{ZhengPami12,HariharanEccv12,TahirPami13,BaktashmotlaghPami14,Chartampilas15,Chartampilas16,YuPami16,DongPami16}, 
face recognition \cite{ChenPami17,LeiPami14,KanPami16}, visual tracking \cite{BalntasPami17,HuPami17}
and other domains.
Used in combination with traditional classifiers, these approaches can effectively deal with
the curse of dimensionality problem by discovering a lower dimensional subspace where class structure is preserved,
thus yielding improved efficiency and classification accuracy \cite{Vapnik98,HouIeeeTnnls15}.

Kernel discriminant analysis (KDA) \cite{Muller01,Mika00}, its subclass extension, KSDA \cite{You11},
and their variants \cite{Baudat00,Gkalelis13NNLS,Gkalelis14},
are among the most powerful DA techniques.
Their major advantage over their linear counterparts
is that they can deal with nonlinear problems more effectively.
Specifically, by carefully selecting an appropriate Mercer kernel \cite{Vapnik98,HofSchSmo08,Mercer_09,JayasumanaPami15},
observations in the input space are implicitly mapped into a new
feature space (usually of much higher-dimensionality), where classes are expected to be linearly separable.
In this space, a linear transformation to a
much lower-dimensional subspace (called discriminant subspace)
is then computed that preserves class separability.
This is usually achieved by solving an optimization problem 
to identify the transformation matrix that maximizes the ratio
of the between- to within-class scatter (or the between- to within-subclass scatter for KSDA)
in the discriminant subspace.
In order to avoid working directly in the feature space (which may be very high- or infinitely-dimensional)
the optimization problem is entirely reformulated in terms of dot products,
which in turn are replaced with kernel function evaluations.
Ultimately, the optimization problem is reformulated to an equivalent
symmetric positive semidefinite (SPSD)
generalized eigenproblem (GEP) and the task is now to identify
the eigenvectors corresponding to the nonzero eigenvalues of the GEP.

The mathematical treatment of symmetric positive definite GEP (SPD GEP)
is one of the most aesthetically pleasing problems in all numerical algebra.
In practice, it is typically solved using techniques that simultaneously
reduce to diagonal form the associated matrices \cite{Golub13}.
However, as we also show in this paper, a major limitation of KDA is that
the kernel matrix associated with the
within-class or -subclass scatter variability is always singular \cite{Muller01,Mika00,Park05J}.
One way to mitigate this problem is to employ a regularization method 
\cite{Muller01,Mika00,Zhang10,Baudat00,Cai11_J}.
For instance, in \cite{Mika00} a small positive regularization constant is added along the diagonal
of the kernel matrix associated with the within-class variability, and subsequently the Cholesky
factorization and the symmetric QR algorithm are applied.
Similarly, in \cite{Baudat00,Cai11_J} the kernel matrix of the centered
training observations is regularized and its eigenvalue decomposition (EVD) or Cholesky factorization is then computed.
Another way to alleviate this limitation is to use a cascade of singular value decompositions (SVDs) \cite{Park05J,Xiong06,Bodesheim13}.
For example, in \cite{Park05J} the kernel matrices associated
with the between- and within-class scatter are factorized and the generalized SVD (GSVD)
is applied on the derived factors.
In \cite{Xiong06}, the null space of the kernel matrix associated with the total scatter is removed,
and in the resulting space, the transformation that performs the simultaneous reduction of the
between- and within-class scatter is computed. This method is called kernel uncorrelated
discriminant analysis (KUDA) due to the fact that the derived transformation yields
uncorrelated features in the discriminant subspace. Using a further processing step
to orthogonalize the columns of the derived transformation matrix,
kernel orthogonal discriminant analysis (KODA)
is also proposed in the same paper. In \cite{Bodesheim13}, a null-space variant of KDA,
called hereafter kernel null discriminant analysis (KNDA), is proposed, that maximizes the between-class scatter in the null space of the within-class scatter matrix (see also \cite{Ye06P,Ye06J}).
In \cite{Ye06P,Ye06J,Xiong06}, it is shown that when the rank of the total scatter matrix equals to the sum of the ranks of
the between- and within-class scatter matrices, then KNDA and KUDA (or KODA for orthogonalized projection vectors) are equivalent.

Despite their success in various domains, the high computational cost of the kernel-based DA methods
described above has prohibited their widespread use in today's
large-scale data problems. Motivated by this limitation, in \cite{Cai11_J} spectral regression kernel
discriminant analysis (SRKDA) is proposed, which efficiently computes the required transformation matrix.
However, SRKDA requires centered data, violating its major
assumption for SPD kernel matrix \cite{Park05J}.
Additionally, the data centering required during both the training and testing phases of SRKDA
increases computational cost and round-off errors,
which have a negative effect in efficiency and classification accuracy.
More importantly, the mathematical framework introduced in \cite{Cai11_J}
is not easily applicable for the acceleration of other DA methods.
To this end, utilizing a new acceleration framework, AKDA and AKSDA are proposed in this paper.
The proposed methods compute efficiently the simultaneous reduction of the associated scatter matrices
using algorithms with excellent numerical stability, and
without posing any prior requirements on data properties.
In summary, the contributions of this paper are:
\begin{itemize}
    \item A novel factorization and simultaneous reduction framework
for discriminant analysis is presented, and, based on
this, efficient variants of KDA and KSDA, namely AKDA and AKSDA, are proposed.
    \item Theoretical links between the proposed methods
and previous kernel-based DA approaches, such as KNDA, KUDA, and KODA, are presented.
    \item An extensive experimental evaluation on various tasks and datasets, ranging from video event to object detection,
is presented, verifying the state-of-the-art performance of the proposed methods 
in both training time and classification accuracy.
\end{itemize}
The rest of the paper is structured as follows. The fundamentals of KDA and KSDA are discussed in Section \ref{S:Fundamentals},
while Section \ref{S:SmltRdct} reviews prior work on simultaneous reduction.
In Sections \ref{S:AKDA} and \ref{S:AKSDA}, the proposed AKDA and AKSDA are presented.
Experimental results and conclusions are provided in Sections \ref{S:exprms} and \ref{S:conclusions}, respectively.

\section{Fundamentals} \label{S:Fundamentals}

Let $\mathbf{X}$ be a training matrix of $C$ disjoint classes,
$H = \sum_{i=1}^C H_i$ disjoint subclasses and $N = \sum_{i=1}^C N_i$ observations
$\mathbf{x}_n$ in the input space $\mathbb{R}^L$, where $H_i$, $N_i$ is the number of
subclasses and observations of the $i$-th class, respectively,
\begin{equation}
\mathbf{X} = [\mathbf{x}_1, \dots, \mathbf{x}_N ] \in
\mathbb{R}^{L \times N}. \label{E:DataMatInpSpc}
\end{equation}
We assume that the observations in $\mathbf{X}$
are sorted in ascending order with respect to their class-subclass label, and
$Y_i$, $Y_{i,j}$ are the sets of $\mathbf{X}$'s column indices
whose observations belong to class $i$  or to subclass $j$ of class $i$, respectively.
KDA \cite{Muller01,Mika00} utilizes a Mercer kernel
$k(\cdot, \cdot)$ \cite{HofSchSmo08,Mercer_09,JayasumanaPami15} associated with a mapping $\bg{\phi}(\cdot)$
\begin{eqnarray}
k(\cdot, \cdot) &:& \mathbb{R}^L \times \mathbb{R}^L \mapsto \mathbb{R}, \label{E:KernelMap}\\
\bg{\phi} (\cdot) &:& \mathbb{R}^L \mapsto \mathbb{R}^F, \label{E:FeaMap} \nonumber
\end{eqnarray}
where, $k_{n,\nu} = k(\mathbf{x}_{n}, \mathbf{x}_{\nu}) = \bg{\phi}_{n}^T \bg{\phi}_{\nu}$,
$\bg{\phi}_n = \bg{\phi} (\mathbf{x}_{n})$ and $\mathbb{R}^F$ is called the feature space.
Given $\mathbf{X}$, it then seeks a linear transformation
$\mathbf{\Gamma} \in \mathbb{R}^{F \times D}$, $D \ll F$, that simultaneously maximizes the
between-class and minimizes the within-class sum of squares in the discriminant subspace $\mathbb{R}^D$.
This simultaneous optimization is commonly approximated by appropriate criteria such as
\cite{Ye06J,Howland04_J,Park08_J}
\begin{equation}
\argmax_{\mathbf{\Gamma}}  \trace( (\mathbf{\Gamma}^T \mathbf{\Sigma}_{w} \mathbf{\Gamma})^{\dagger} 
\mathbf{\Gamma}^T \mathbf{\Sigma}_{b} \mathbf{\Gamma} ), \label{E:kdaTrCritFea}
\end{equation}
which usually require the identification of the eigenvector matrix
$\mathbf{\Gamma}$ satisfying the following GEP
\begin{equation}
\mathbf{\Sigma}_b \mathbf{\Gamma} =  \mathbf{\Sigma}_w \mathbf{\Gamma}
\mathbf{\Lambda}, \label{E:kdaGepFea}
\end{equation}
where, $\trace(\mathbf{A})$, $\mathbf{A}^T$ and $\mathbf{A}^\dagger$
denote trace, transpose and pseudoinverse of any matrix $\mathbf{A}$, respectively,
$\mathbf{\Sigma}_b$, $\mathbf{\Sigma}_w$
are the between- and within-class scatter matrices,
\begin{eqnarray}
\mathbf{\Sigma}_{b} &=& \sum_{i=1}^{C} N_{i} (\bg{\mu}_{i} -
\bg{\mu}) (\bg{\mu}_{i} - \bg{\mu})^{T}, \label{E:SigmbFea} \\
\mathbf{\Sigma}_{w} &=& \sum_{i=1}^{C} \sum_{n \in Y_{i}}
(\bg{\phi}_n - \bg{\mu}_{i}) (\bg{\phi}_n -
\bg{\mu}_{i})^{T}, \label{E:SigmwFea}
\end{eqnarray}
$\bg{\mu}_{i} = \frac{1}{N_i} {\sum}_{n \in Y_{i}} \bg{\phi}_n$
is the estimated sample mean of class $i$,
$\mathbf{\Lambda} = \diag(\lambda_1, \dots, \lambda_D), \lambda_1 \geq \dots  \geq \lambda_D > 0$,
$\lambda_d$ is the $d$th largest generalized eigenvalue
and $D = \rank(\mathbf{\Sigma}_b)$.
That is, $(\mathbf{\Gamma}, \mathbf{\Lambda})$ contain the NZEP
of the matrix pencil $(\mathbf{\Sigma}_b, \mathbf{\Sigma}_w)$.
The implicit mapping $\bg{\phi} (\cdot)$ may be unknown,
or the dimensionality of the feature space
$F$ may be very high or even infinite.
In order to avoid working directly in $\mathbb{R}^F$,
the GEP in (\ref{E:kdaGepFea})
is typically reformulated in terms of dot products,
which in turn are replaced with kernel function evaluations
as explained in the following.
Starting from $\mathbf{\Gamma}$, its solution space
in $\mathbb{R}^F$ is restricted to $\spanvs(\mathbf{\Phi})$ \cite{Mika00,Park05J}
where $\mathbf{\Phi}$ denotes the training matrix in the feature space
\begin{equation}
\mathbf{\Phi} = [\bg{\phi}_1, \dots, \bg{\phi}_N ] \in
\mathbb{R}^{F \times N}. \label{E:DataMatFeaSpc} \nonumber
\end{equation}
This allows us to express each column of
$\mathbf{\Gamma}$ as a linear combination of the mapped training data,
$\mathbf{\Gamma} = \mathbf{\Phi} \mathbf{\Psi}$,
where $\mathbf{\Psi} \in \mathbb{R}^{N \times D}$ contains the
expansion coefficients. Substituting the above expression of
$\mathbf{\Gamma}$ into (\ref{E:kdaGepFea}) and pre-multiplying with $\mathbf{\Phi}^T$
we get the following equivalent GEP
\begin{equation}
\mathbf{S}_b \mathbf{\Psi} =  \mathbf{S}_w \mathbf{\Psi}
\mathbf{\Lambda}, \label{E:kdaGepKer} \nonumber
\end{equation}
where, $\mathbf{S}_b = \mathbf{\Phi}^T \mathbf{\Sigma}_b \mathbf{\Phi}$,
$\mathbf{S}_w = \mathbf{\Phi}^T \mathbf{\Sigma}_w \mathbf{\Phi}$,
are the kernel matrices associated with the between- and within-class variability,
fully expressed in terms of dot products
\begin{eqnarray}
\mathbf{S}_b &=& \sum_{i=1}^{C} N_i ( \mathbf{K}_i \mathbf{\tilde{1}}_{N_i} - \mathbf{K} \mathbf{\tilde{1}}_{N} ) 
( \mathbf{K}_i \mathbf{\tilde{1}}_{N_i} - \mathbf{K} \mathbf{\tilde{1}}_{N} ) ^T, \label{E:SbKer} \\
\mathbf{S}_w &=& \sum_{i=1}^{C} \sum_{n \in Y_i}^{C} ( \mathbf{k}_n - \mathbf{K}_i \mathbf{\tilde{1}}_{N_i} )
 ( \mathbf{k}_n - \mathbf{K}_i \mathbf{\tilde{1}}_{N_i} )^T, \label{E:SwKer}
\end{eqnarray}
$\mathbf{K}$ is the kernel (or Gram) matrix of the training observations,
\begin{equation}
\mathbf{K} = \mathbf{\Phi}^T \mathbf{\Phi} = [\mathbf{k}_1, \dots, \mathbf{k}_N], \label{E:KerMat}
\end{equation}
$\mathbf{k}_n = [k(\mathbf{x}_1, \mathbf{x}_n) , \dots, k(\mathbf{x}_N, \mathbf{x}_n) ]^T$
is the kernel vector associated with the $n$th training observation,
$\mathbf{K}_i = \mathbf{\Phi}^T \mathbf{\Phi}_i$ and $\mathbf{\Phi}_i$ are the 
block matrices of $\mathbf{K}$ and $\mathbf{\Phi}$, respectively,
whose columns correspond to the training observations of the $i$th class,
$\mathbf{\tilde{1}}_{N_i} = \frac{1}{N_i} \mathbf{1}_{N_i}$,
$\mathbf{\tilde{1}}_{N} = \frac{1}{N} \mathbf{1}_{N}$, and
$\mathbf{1}_{N}, \mathbf{1}_{N_i}$ are all-one vectors with dimension $N$ and $N_i$, respectively.
That is, instead of solving (\ref{E:kdaGepFea}),
we now need to identify the NZEP $(\mathbf{\Psi}, \mathbf{\Lambda})$
of the matrix pencil $(\mathbf{S}_b, \mathbf{S}_w)$.
In practice, this is accomplished
using techniques that compute the simultaneous reduction of $\mathbf{S}_b$ and $\mathbf{S}_w$
\begin{equation}
\begin{array}{l}
 \mathbf{\Psi}^T \mathbf{S}_b \mathbf{\Psi} =  \mathbf{\Delta} = \diag(\delta_1, \dots, \delta_D), \\
 \mathbf{\Psi}^T \mathbf{S}_w \mathbf{\Psi} =  \mathbf{\Upsilon} = \diag(\upsilon_1, \dots, \upsilon_D),
\end{array} \label{E:smltRdct}
\end{equation}
where $\delta_1 / \upsilon_1 \ge, \dots, \ge \delta_D / \upsilon_D > 0$, 
and for $\upsilon_d = 0$, $\frac{\delta_d}{\upsilon_d}$ is set to $+\infty$,
i.e. the limit of $\frac{\delta_d}{\upsilon_d}$
as $\upsilon_d$ approaches zero from the right is considered \cite{Park05J,Zhang10,Golub13}.
The projection $\mathbf{z} \in \mathbb{R}^{D}$ of a test vector $\mathbf{x} \in \mathbb{R}^{L}$
can be then computed using
$\mathbf{z} = \mathbf{\Gamma}^T \bg{\phi} =
\mathbf{\Psi}^T \mathbf{\Phi}^T \bg{\phi} = \mathbf{\Psi}^T \mathbf{k}$,
where
\begin{equation}
 \mathbf{k} = [k(\mathbf{x}_1, \mathbf{x}), \dots, k(\mathbf{x}_N, \mathbf{x}) ]^T. \label{E:kernelRpr}
\end{equation}

A fundamental assumption of KDA is that classes are distributed unimodally in the feature space.
However, in certain applications, classes may be multimodal or artificially divided to subclasses.
To this end, KSDA \cite{You11,Gkalelis13NNLS,Gkalelis14}
uses a more flexible criterion, allowing the incorporation of subclass
information into the optimization problem,
\begin{equation}
\argmax_{\mathbf{G}}  \trace( (\mathbf{G}^T \mathbf{\Sigma}_{ws} \mathbf{G})^{\dagger} 
\mathbf{G}^T \mathbf{\Sigma}_{bs} \mathbf{G} ), \label{E:ksdaTrCritFea}
\end{equation}
where, $\mathbf{G} \in \mathbb{R}^{F \times \mathcal{D}}$, $\mathcal{D} \ll F$,
is the transformation matrix to be identified, $\mathbf{\Sigma}_{bs}$, $\mathbf{\Sigma}_{ws}$,
are the between- and within-subclass scatter matrices,
\begin{eqnarray}
\mathbf{\Sigma}_{bs} &=& \frac{1}{N} \sum_{i=1}^{C-1} \sum_{j=1}^{H_i}
\sum_{k=i+1}^{C} \sum_{l=1}^{H_k} N_{i,j} N_{k,l} (\bg{\mu}_{i,j} -
\bg{\mu}_{k,l}) \nonumber \\ & & \times (\bg{\mu}_{i,j} -
\bg{\mu}_{k,l})^{T}, \label{E:SigmbsFea} \\
\mathbf{\Sigma}_{ws} &=& \sum_{i=1}^{C} \sum_{j=1}^{H_i} \sum_{n \in Y_{i,j}}
(\bg{\phi}_n - \bg{\mu}_{i,j}) (\bg{\phi}_n - \bg{\mu}_{i,j})^{T}, \label{E:SigmwsFea}
\end{eqnarray}
and $N_{i,j}$, $\bg{\mu}_{i,j} = \frac{1}{N_{i,j}} {\sum}_{n \in
Y_{i,j}} \bg{\phi}_n$ are the number of observations and
estimated sample mean of subclass $(i,j)$, respectively.
Equivalently, the following GEP may be solved
\begin{equation}
\mathbf{\Sigma}_{bs} \mathbf{G} =  \mathbf{\Sigma}_{ws} \mathbf{G}
\mathbf{\breve{\Lambda}}, \label{E:ksdaGepFea}
\end{equation}
where $\mathbf{\breve{\Lambda}} \in \mathbb{R}^{\mathcal{D} \times \mathcal{D}}$ is the diagonal matrix
containing the positive eigenvalues of the GEP, sorted in descending order.
The transformation matrix $\mathbf{G}$ is further expressed
as the linear combination of the mapped training data,
$\mathbf{G} = \mathbf{\Phi} \mathbf{W}$, where,
$\mathbf{W} \in \mathbb{R}^{N \times \mathcal{D}}$ is the respective coefficient matrix.
Pre-multiplying (\ref{E:ksdaGepFea}) with $\mathbf{\Phi}^T$ and using the expression of $\mathbf{G}$ above, 
the GEP in (\ref{E:ksdaGepFea}) is transformed to
\begin{equation}
\mathbf{S}_{bs} \mathbf{W} =  \mathbf{S}_{ws} \mathbf{W}
\mathbf{\breve{\Lambda}}, \label{E:aksdaGepKer}
\end{equation}
where, $\mathbf{S}_{bs} = \mathbf{\Phi}^T \mathbf{\Sigma}_{bs} \mathbf{\Phi}$,
$\mathbf{S}_{ws} = \mathbf{\Phi}^T \mathbf{\Sigma}_{ws} \mathbf{\Phi}$,
are the kernel matrices associated with the between- and within-subclass scatter matrices,
and can be fully expressed in terms of kernel function evaluations,
\begin{eqnarray}
\mathbf{S}_{bs} &=& \sum_{i=1}^{C-1} \sum_{j=1}^{H_i}
\sum_{k=i+1}^{C} \sum_{l=1}^{H_k} \frac{ N_{i,j} N_{k,l} }{N}  ( \bg{\eta}_{i,j} - \bg{\eta}_{k,l} ) \nonumber \\
&& \times
   ( \bg{\eta}_{i,j} - \bg{\eta}_{k,l}  ) ^T, \label{E:SbsKer} \\
\mathbf{S}_{ws} &=& \sum_{i=1}^{C} \sum_{j=1}^{H_i}
 \sum_{n \in Y_{i,j}}^{C} ( \mathbf{k}_n - \bg{\eta}_{i,j} )   ( \mathbf{k}_n - \bg{\eta}_{i,j} )^T, \label{E:SwsKer}
\end{eqnarray}
$\bg{\eta}_{i,j} = \mathbf{K}_{i,j} \mathbf{\tilde{1}}_{N_{i,j}}$,
$\mathbf{\tilde{1}}_{N_{i,j}} = \frac{1}{N_{i,j}} \mathbf{1}_{N_{i,j}}$, and
$\mathbf{K}_{i,j} = \mathbf{\Phi}^T \mathbf{\Phi}_{i,j}$, $\mathbf{\Phi}_{i,j}$ are the 
block matrices of $\mathbf{K}$ and $\mathbf{\Phi}$, respectively,
whose columns correspond to the training observations of subclass $(i,j)$.
As in KDA (\ref{E:smltRdct}), the optimization problem in (\ref{E:aksdaGepKer})
is in practice solved by computing the coefficient matrix $\mathbf{W}$
that performs the simultaneous reduction of $\mathbf{S}_{bs}$ and $\mathbf{S}_{ws}$.

We should note that $\mathbf{\Sigma}_{w}$, $\mathbf{\Sigma}_{ws}$
are often replaced by the total scatter matrix,
\begin{equation}
\mathbf{\Sigma}_{t} = \sum_{n=1}^{N} (
\bg{\phi}_{n} - \bg{\mu}) ( \bg{\phi}_{n} - \bg{\mu} )^T, \label{E:SigmtFea}
\end{equation}
where $\bg{\mu} = \frac{1}{N} \sum_{n=1}^{N} \bg{\phi}_{n}$ is the estimated total sample mean.
In this case, $\mathbf{S}_{w}$ or $\mathbf{S}_{ws}$ are substituted by the respective kernel matrix
\begin{equation}
\mathbf{S}_{t} = \mathbf{\Phi}^T \mathbf{\Sigma}_{t} \mathbf{\Phi} = \sum_{n=1}^{N} ( \mathbf{k}_{n} - \mathbf{K} \mathbf{\tilde{1}}_N ) ( \mathbf{k}_{n} - \mathbf{K} \mathbf{\tilde{1}}_N )^T. \label{E:StKer}
\end{equation}

\section{Simultaneous reduction}  \label{S:SmltRdct}

As explained in the previous section, the simultaneous reduction of
$\mathbf{S}_b$, $\mathbf{S}_w$ (or $\mathbf{S}_t$) for KDA, and
$\mathbf{S}_{bs}$, $\mathbf{S}_{ws}$ (or $\mathbf{S}_t$) for KSDA,
plays an essential role in the computation of the respective transformation matrix.
Noting that $\mathbf{S}_w$, $\mathbf{S}_{ws}$ and $\mathbf{S}_t$ are SPSD,
the way we achieve the simultaneous reductions above strongly relates
with the way we deal with the singularity of these matrices.
In the following, the different approaches for mitigating the singularity problem
are categorized to regularization- and SVD-based.

\subsection{Regularization-based approaches} \label{SS:RegBased}

One way to deal with the singularity of  $\mathbf{S}_w$, $\mathbf{S}_{ws}$ or $\mathbf{S}_t$,
is using a ridge-type regularization operator \cite{mika99,Muller01,Mika00,You11}.
For instance, in \cite{mika99}, the simultaneous reduction of $\mathbf{S}_b$ and
$\mathbf{S}_w \leftarrow \mathbf{S}_w + \epsilon \mathbf{I}_N$ is computed
using the Cholesky factorization of $\mathbf{S}_w$
and the symmetric QR algorithm, where $\mathbf{I}_N$ is the $N \times N$ identity matrix,
and $\epsilon > 0$ is a small regularization constant\footnote{Note
that the diagonalization of $\mathbf{S}_w^{-1} \mathbf{S}_b$
is not the preferred approach as this matrix is not symmetric.}.
Another set of techniques perform zero mean normalization
and subsequently apply regularization on the resulting kernel matrix
 \cite{Baudat00,Gkalelis14,Cai11_J}. For instance, GDA \cite{Baudat00} computes the
simultaneous reduction of the kernel matrices expressing the between-class and total variability of the centered training data,
$\mathbf{\bar{S}}_b = \mathbf{\bar{K}} \mathbf{\bar{C}} \mathbf{\bar{K}}$,
$\mathbf{\bar{S}}_t = \mathbf{\bar{K}} \mathbf{\bar{K}}$, respectively, where
$\mathbf{\bar{C}}$ is the block diagonal matrix defined as
$\mathbf{\bar{C}} = \diag(\mathbf{\bar{C}}_1, \dots, \mathbf{\bar{C}}_C)$,
$\mathbf{\bar{C}}_i = \frac{1}{N_i} \mathbf{J}_{N_i}$ is the $i$th block in the diagonal of $\mathbf{\bar{C}}$,
$\mathbf{J}_{N_i}$ is the $N_i \times N_i$
all-one matrix, and $\mathbf{\bar{K}} $ is the kernel matrix of the centered data,
\begin{eqnarray}
 \mathbf{\bar{K}} &=& \mathbf{K} - \frac{1}{N} \mathbf{K}
 \mathbf{J}_{N} - \frac{1}{N} \mathbf{J}_{N} \mathbf{K} 
 + \frac{1}{N^2} \mathbf{J}_{N} \mathbf{K} \mathbf{J}_{N}. \label{E:GramCntrd}
\end{eqnarray}
Due to data centering, $\mathbf{\bar{K}}$ is always singular and usually replaced by
$\mathbf{\bar{K}} \leftarrow \mathbf{\bar{K}} + \epsilon \mathbf{I}_N$ \cite{mika99,Park05J}.
Similar to GDA, SRKDA \cite{Cai11_J} performs the simultaneous reduction of
$\mathbf{\bar{S}}_b$ and $\mathbf{\bar{S}}_t$, however,
using a different approach that provides a significant computational gain.
In particular, the coefficient matrix $\mathbf{\bar{\Psi}}$
that performs the simultaneous reduction above
is computed by solving the following
linear matrix system $\mathbf{\bar{K}} \mathbf{\bar{\Psi}} = \mathbf{\bar{\Theta}}$,
where $\mathbf{\bar{\Theta}}$'s columns are the eigenvectors of
$\mathbf{\bar{C}}$ corresponding to the nonzero eigenvalues.
The speed up is achieved from the computation of $\mathbf{\bar{\Theta}}$,
which can be performed very efficiently exploiting certain properties of 
$\mathbf{\bar{C}}$ and the Gram-Schmidt process.
Note that GDA and SRKDA require zero mean normalization
during both training and testing.
For instance, the projection $\mathbf{\bar{z}}$ of a test observation $\mathbf{x}$ is computed using
\begin{equation}
\bar{\mathbf{z}} = \mathbf{\Psi}^T (\mathbf{k} - \frac{1}{N} \mathbf{K} \mathbf{1}_N),
\label{E:prjTstObsCntrd}
\end{equation}
where $\mathbf{k}$ is defined in (\ref{E:kernelRpr}).
This can have a negative effect in the performance of these methods
due to the additional computational cost and round-off errors
introduced with (\ref{E:GramCntrd}) and (\ref{E:prjTstObsCntrd}).

\subsection{SVD-based approaches}

In contrary to using a regularization step, a cascade of SVD steps 
is used by several methods to bypass the singularity problem.
For instance, KDA/GSVD \cite{Park05J,Park08_J},
based on the factorizations 
$\mathbf{S}_b = \mathbf{K}_b \mathbf{K}_b^T$,
$\mathbf{S}_w = \mathbf{K}_w \mathbf{K}_w^T$,
$\mathbf{S}_t = \mathbf{K}_t \mathbf{K}_t^T$,
and the equality
$\mathbf{S}_t = \mathbf{S}_b + \mathbf{S}_w$,
performs the GSVD of $[ \mathbf{K}_b, \mathbf{K}_w ]^T$
to derive the required transformation matrix, where
\begin{equation}
\begin{array}{l}
\mathbf{K}_b = \mathbf{\Phi}^T [ \sqrt{N_1} (\bg{\mu}_1 - \bg{\mu}), \dots, \sqrt{N_C} (\bg{\mu}_C - \bg{\mu})] \\
 \quad \; = [\sqrt{N_1} (\mathbf{K}_1 \mathbf{\tilde{1}}_{N_1} - \mathbf{K} \mathbf{\tilde{1}}_N), \dots, 
  \sqrt{N_C} (\mathbf{K}_C \mathbf{\tilde{1}}_{N_C} - \mathbf{K} \mathbf{\tilde{1}}_N)], \\
\mathbf{K}_w = \mathbf{\Phi}^T [\mathbf{\Phi}_1 - \bg{\mu}_1 \mathbf{1}_{N_1}^T, \dots,
\mathbf{\Phi}_C - \mathbf{m}_C \mathbf{1}_{N_C}^T] \\
 \quad \; = [ \mathbf{K}_1( \mathbf{I}_{N_1} - \frac{1}{N_1} \mathbf{J}_{N_1}), \dots,
\mathbf{K}_C ( \mathbf{I}_{N_C} - \frac{1}{N_1} \mathbf{J}_{N_C} ) ], \\
 \mathbf{K}_t = \mathbf{\Phi}^T (\mathbf{\Phi} - \bg{\mu} \mathbf{1}_{N}^T)
= \mathbf{K} ( \mathbf{I}_N - \frac{1}{N} \mathbf{J}_N ).
\end{array} \nonumber
\end{equation}
Similarly, KUDA \cite{Xiong06} performs the SVD of $\mathbf{K}_t$ and subsequently of
$\mathbf{K}_b$ in the range space of $\mathbf{S}_t$.
The computed transformation is a whitening transform
yielding uncorrelated features in the derived subspace.
A further processing step is applied in KODA \cite{Xiong06} to orthogonalize the columns of the
transformation matrix retrieved using KUDA.
Finally, KNDA-based approaches \cite{Ye06P,Ye06J,Bodesheim13},
perform the SVD of $\mathbf{K}_w$ and subsequently,
the SVD of $\mathbf{K}_b$ in the null space of $\mathbf{S}_w$.
Ultimately, the transformation and coefficient matrices,
$\mathbf{\tilde{\Gamma}}$, $\mathbf{\tilde{\Psi}}$, respectively, retrieved with either of the methods above
satisfy the following simultaneous reduction
\begin{eqnarray}
 \mathbf{\tilde{\Gamma}}^T \mathbf{\Sigma}_b \mathbf{\tilde{\Gamma}}
 &=& \mathbf{\tilde{\Psi}}^T \mathbf{S}_b \mathbf{\tilde{\Psi}} \nonumber = \mathbf{\tilde{\Delta}}, \nonumber \\
 \mathbf{\tilde{\Gamma}}^T \mathbf{\Sigma}_w \mathbf{\tilde{\Gamma}}
 &=& \mathbf{\tilde{\Psi}}^T \mathbf{S}_w \mathbf{\tilde{\Psi}} \nonumber = \mathbf{\tilde{\Upsilon}}, \nonumber \\
 \mathbf{\tilde{\Gamma}}^T \mathbf{\Sigma}_t \mathbf{\tilde{\Gamma}}
 &=& \mathbf{\tilde{\Psi}}^T \mathbf{S}_t \mathbf{\tilde{\Psi}} \nonumber = \mathbf{\tilde{\Delta}} + \mathbf{\tilde{\Upsilon}},  \label{E:smltRdctKdaGsda} \nonumber
\end{eqnarray}
where, $\mathbf{\tilde{\Delta}} = \diag( \tilde{\delta}_1, \dots, \tilde{\delta}_D)$,
$\mathbf{\tilde{\Upsilon}} = \diag( \tilde{\upsilon}_1, \dots, \tilde{\upsilon}_D)$,
$\tilde{\delta}_1 \ge \dots \ge \tilde{\delta}_D > 0$,
$\tilde{\upsilon}_D \ge \dots \ge \tilde{\upsilon}_{\rho} = \dots = \tilde{\upsilon}_1 = 0$,
$0 < \rho < D$, and,
$\mathbf{\tilde{\Delta}} + \mathbf{\tilde{\Upsilon}} = \mathbf{I}_D$ for KUDA,
$\mathbf{\tilde{\Gamma}}^T \mathbf{\tilde{\Gamma}} = \mathbf{\tilde{\Psi}}^T \mathbf{K} \mathbf{\tilde{\Psi}}
= \mathbf{I}_D$ for KODA,
and $\mathbf{\tilde{\Delta}} = \mathbf{I}_D$, $\mathbf{\tilde{\Upsilon}} = \mathbf{0}_D$
for KNDA, where $\mathbf{0}_D$ is the $D \times D$ all-zero matrix.
Note that the integer $\rho$ is always larger than zero due to the fact that $\mathbf{S}_w$ is SPSD.
It can be shown that under the following condition
\begin{equation}
 \rank( \mathbf{\Sigma}_t) = \rank( \mathbf{\Sigma}_b ) + \rank( \mathbf{\Sigma}_w ),
\label{C:rnkCndt}
\end{equation}
KNDA, is equivalent to KUDA, or KODA depending on the properties of the transformation matrix
used \cite{Ye06P,Ye06J,Bodesheim13}.
This condition holds for linearly independent training observations in the feature space, e.g.,
when strictly positive definite kernels are used \cite{HofSchSmo08},
or the linear kernel is utilized and training observations are linearly independent in the input space.
The latter is almost always the case in problems involving high-dimensional undersampled data.

\section{AKDA} \label{S:AKDA}

Given the data matrix $\mathbf{X}$
and class index sets $Y_i$  (\ref{E:DataMatInpSpc}), AKDA computes the transformation matrix
 $\mathbf{\Gamma} \in \mathbb{R}^{F \times D}$  satisfying (\ref{E:kdaGepFea})
and equivalently (\ref{E:kdaTrCritFea}).

\subsection{Factorization}

The scatter matrices $\mathbf{\Sigma}_{b}$ (\ref{E:SigmbFea}), $\mathbf{\Sigma}_{w}$ (\ref{E:SigmwFea})
and $\mathbf{\Sigma}_{t}$ (\ref{E:SigmtFea}), can be expressed as
\begin{eqnarray}
\mathbf{\Sigma}_{b} &=& \sum_{i=1}^{C} N_i \bg{\mu}_i \bg{\mu}_i^T -
\frac{1}{N} \sum_{i=1}^{C} \sum_{j=1}^{C} N_i N_j 
\bg{\mu}_i \bg{\mu}_j^T \nonumber \\
&=& \mathbf{M}_C ( \mathbf{N}_C  - \frac{1}{N} \mathbf{\tilde{N}}_C ) \mathbf{M}_C^T 
\; = \;  \mathbf{\Phi} \mathbf{C}_b \mathbf{\Phi}^T, \label{E:SigmbFac}\\
\mathbf{\Sigma}_{w} &=& \sum_{i=1}^{C} \sum_{n \in Y_i} \bg{\phi}_{n}
\bg{\phi}_{n}^T - \sum_{i=1}^{C} N_i \bg{\mu}_{i}
\bg{\mu}_{i}^T \nonumber \\
 &=& \mathbf{\Phi} \mathbf{\Phi}^T - \mathbf{M}_C \mathbf{N}_C \mathbf{M}_C^T
 \; = \;   \mathbf{\Phi} \mathbf{C}_w \mathbf{\Phi}^T, \label{E:SigmwFac}\\
\mathbf{\Sigma}_{t} &=& \sum_{n=1}^{N} \bg{\phi}_{n} \bg{\phi}_{n}^T
- \frac{1}{N} \sum_{n=1}^{N} \sum_{\nu=1}^{N} \bg{\phi}_{n} \bg{\phi}_{\nu}^T
 \; = \;   \mathbf{\Phi} \mathbf{C}_t \mathbf{\Phi}^T, \;\;\; \label{E:SigmtFac}
\end{eqnarray}
where, $\mathbf{M}_C$ is the matrix whose columns are the class means,
\begin{equation}
 \mathbf{M}_C =  [\bg{\mu}_{1}, \dots, \bg{\mu}_{C}] = \mathbf{\Phi} \mathbf{R}_C \mathbf{N}_C^{-1}, \label{E:Mc}
\end{equation}
$\mathbf{n}_C$, $\mathbf{N}_C$,
are the so-called class strength vector and matrix, respectively,
\begin{eqnarray}
 \mathbf{n}_C &=& [  N_{1}, \dots, N_{C} ]^T, \label{E:classStrengthVec} \nonumber \\
 \mathbf{N}_C &=& \diag( N_{1}, \dots, N_{C} ), \label{E:classStrengthMat} 
\end{eqnarray}
$\mathbf{\tilde{N}}_C$ is the matrix resulting from the outer product of $\mathbf{n}_C$ with itself,
$\mathbf{\tilde{N}}_C = \mathbf{n}_C \mathbf{n}_C^T$,
$ \mathbf{R}_C \in \mathbb{R}^{N \times C}$ is the class indicator matrix whose element $[\mathbf{R}_C]_{n,i}$
corresponding to the $n$th observation and $i$th class is one if $n \in Y_{i}$ and zero otherwise,
\begin{eqnarray}
 \mathbf{C}_{b} &=& \mathbf{R}_C \mathbf{N}_C^{-1/2}
 \mathbf{O}_b \mathbf{N}_C^{-1/2}  \mathbf{R}_C^T \label{E:Cb} \\ 
 \mathbf{C}_{w} &=& \mathbf{I}_N - \mathbf{R}_C \mathbf{N}_C^{-1} \mathbf{R}_C^T, \label{E:Cw} \nonumber\\
 \mathbf{C}_{t} &=& \mathbf{I}_N - \textstyle \frac{1}{N} \mathbf{J}_N \label{E:Bt} \nonumber
\end{eqnarray}
are the so-called central factor matrices of the between-class, within-class and total scatter
matrix, respectively, and, $\mathbf{N}_C^{1/2} = \diag(\sqrt{N_1}, \dots, \sqrt{N_C})$.
In (\ref{E:Cb}), $\mathbf{O}_{b}$ is the so-called core matrix
of the between-class scatter matrix defined as
\begin{equation}
 \mathbf{O}_{b} = \mathbf{N}_C^{-1/2} ( \mathbf{N}_C  - \frac{1}{N} \mathbf{\tilde{N}}_C )
 \mathbf{N}_C^{-1/2} 
= \mathbf{I}_C - \mathbf{\dot{N}}_C, \label{E:Ob}
\end{equation}
where $\mathbf{\dot{N}}_C = \frac{  \mathbf{\dot{n}}_C \mathbf{\dot{n}}_C^T }{ \mathbf{\dot{n}}_C^T \mathbf{\dot{n}}_C }$
and $\mathbf{\dot{n}}_C = [\sqrt{N_1}, \dots, \sqrt{N_C} ]^T$.
Using (\ref{E:SigmbFac}), (\ref{E:SigmwFac}) and (\ref{E:SigmtFac}),
$\mathbf{S}_b$  (\ref{E:SbKer}),
$\mathbf{S}_w$ (\ref{E:SwKer}) and $\mathbf{S}_t$ (\ref{E:StKer})
are similarly expressed as
\begin{equation}
 \mathbf{S}_b = \mathbf{K} \mathbf{C}_b \mathbf{K}, \;\;
 \mathbf{S}_w = \mathbf{K} \mathbf{C}_w \mathbf{K}, \;\;
 \mathbf{S}_t = \mathbf{K} \mathbf{C}_t \mathbf{K}. \nonumber
\end{equation}
The schematic representation of $\mathbf{S}_b$'s factorization is depicted in Figure \ref{fig:akdaVis}.


\subsection{Properties of the factorization}

It is trivial to show that the core and central factor matrices are symmetric.
In the following, the Lemmas described below are necessary.

\begin{lemma} \label{L:prodMatRnk}
Consider the matrices $\mathbf{A} \in \mathbb{R}^{P \times Q}$,
$\mathbf{B} \in \mathbb{R}^{P \times P}$, $P \geq Q$. If 
$\rank( \mathbf{A} ) = Q$ then $\rank(\mathbf{A}
\mathbf{B} \mathbf{A}^T) = \rank(\mathbf{B})$.
\end{lemma}
\textit{Proof}. See \cite{Heinz09}.

\begin{lemma} \label{L:sumMatRnk}
Let $\mathbf{A}$, $\mathbf{B} \in \mathbb{R}^{P \times P}$.
If $\mathbf{B}^T \mathbf{A} = \mathbf{A}^T \mathbf{B} = \mathbf{0}_P$
then $\rank(\mathbf{A} + \mathbf{B}) = \rank(\mathbf{A}) +
\rank(\mathbf{B})$.
\end{lemma}
\textit{Proof}. See \cite{Meyer69,Marsaglia72}.

\begin{lemma} \label{L:idempotRnkRelat}
Let $\mathbf{A} \in \mathbb{R}^{P \times P}$ be symmetric
idempotent with $\rank(\mathbf{A}) = Q$,
and $\mathbf{B} = \mathbf{I}_P - \mathbf{A}$.
Then, $\mathbf{A}$, $\mathbf{B}$ are mutually orthogonal,
and $\mathbf{B}$ is also symmetric idempotent with $\rank(\mathbf{B}) = P - Q$.
\end{lemma}
\textit{Proof}. It is trivial to show that $\mathbf{B}$ is
symmetric. For the idempotency we have,
$\mathbf{B} \mathbf{B} = \mathbf{I}_P - 2 \mathbf{A} +
\mathbf{A} \mathbf{A} = \mathbf{B}$.
It is also easy to show that $\mathbf{B}$ and $\mathbf{A}$
are mutually orthogonal and their sum is the identity matrix
\begin{eqnarray}
\mathbf{B} \mathbf{A} &=& \mathbf{A} \mathbf{B} = \mathbf{A} -
\mathbf{A} \mathbf{A} = \mathbf{0}_P, \nonumber \\
\mathbf{B} + \mathbf{A} &=& \mathbf{I}_P - \mathbf{A} + \mathbf{A} =
\mathbf{I}_P, \nonumber
\end{eqnarray}
Then, using Lemma \ref{L:sumMatRnk} we arrive to
\begin{equation}
\rank(\mathbf{B}) =  \rank(\mathbf{B} + \mathbf{A}) -
\rank(\mathbf{A}) = P -  Q. \nonumber
\end{equation}
A special case for the above Lemma is when $\mathbf{B}$ is a
rank one update of the identity matrix, 
$\mathbf{B} = \mathbf{I}_{P} - \frac{\mathbf{v}
\mathbf{v}^T}{\mathbf{v}^T \mathbf{v}}$,
where $\mathbf{v} \in \mathbb{R}^P$. Then, it is easy to
show that $\rank(\mathbf{B}) = P -1$. Moreover, $\frac{\mathbf{v}
\mathbf{v}^T}{\mathbf{v}^T \mathbf{v}}$ is an orthogonal projection
to $\spanvs(\mathbf{v})$, and, thus,
$\rangesp(\mathbf{B}) = \spanvs(\mathbf{v})^\perp$,
where $\mathcal{S}^\perp$ denotes the orthogonal complement of the subspace $\mathcal{S}$.

Turning back to the properties of the factorization,
it is easy to show that the matrices $\mathbf{\dot{N}}_C$ (\ref{E:Ob}) and
$\frac{1}{N} \mathbf{J}_N$ are idempotent.
Considering that $\mathbf{R}_C^T \mathbf{R}_C = \mathbf{N}_C$,
the same is true for $\mathbf{R}_C \mathbf{N}_C^{-1} \mathbf{R}_C^T$.
Using Lemma \ref{L:idempotRnkRelat}, and the idempotency of the above matrices,
we can easily show that $\mathbf{O}_b$, $\mathbf{C}_b$, $\mathbf{C}_w$,
and $\mathbf{C}_t$ are also idempotent. Due to the fact that they are real symmetric idempotent
they are also SPSD. Using Lemma \ref{L:idempotRnkRelat} and the fact that
$\mathbf{\dot{N}}_C$ in (\ref{E:Ob}) is a rank-one update of the identity matrix we can easily show that 
\begin{eqnarray}
\rank(\mathbf{O}_b) &=& C - 1, \label{E:ObRnk}\\
\rangesp(\mathbf{O}_b) &=& \spanvs(\mathbf{\dot{n}}_C)^\perp. \label{E:ObRng}
\end{eqnarray}
By applying two times Lemma
\ref{L:prodMatRnk} and considering that both $\mathbf{I}_{C}$,
$\mathbf{N}_C$ are of full column rank, we can write
\begin{equation}
\rank(\mathbf{C}_{b}) = \rank(\mathbf{N}_{C}^{-1/2}
\mathbf{O}_{b} \mathbf{N}_{C}^{-1/2}) =
\rank(\mathbf{O}_{b}) = C - 1.  \label{E:rnkCb}\\
\end{equation}
Using Lemma \ref{L:prodMatRnk} and the fact that
$\mathbf{R}_C$ is of full column rank we can show that
$\rank(\mathbf{R}_C \mathbf{N}_C^{-1} \mathbf{R}_C^T) =
\rank(\mathbf{N}_C^{-1}) = C$. Then, using Lemma \ref{L:idempotRnkRelat}
and the idempotency of $\mathbf{R}_C \mathbf{N}_C^{-1} \mathbf{R}_C^T$
we arrive to
\begin{equation}
\rank( \mathbf{C}_{w}) = N - C. \label{E:rnkCw}
\end{equation}
Noting that
$\frac{1}{N} \mathbf{J}_N = \frac{\mathbf{1}_N
\mathbf{1}_N^T}{\mathbf{1}_N^T \mathbf{1}_N}$,
we conclude that $\mathbf{C}_t$ is a rank-one
update of the identity matrix, and using Lemma 
\ref{L:idempotRnkRelat} we can write
\begin{equation}
\rank( \mathbf{C}_t) = N - 1. \label{E:rnkCt}
\end{equation}
It is also easy to show that $\mathbf{C}_{t} = \mathbf{C}_{b} + \mathbf{C}_{w}$.
Using (\ref{E:rnkCb}), (\ref{E:rnkCw}), (\ref{E:rnkCt}),
we arrive to the following inequalities
\begin{eqnarray}
 \rank( \mathbf{S}_b ) &\le& C - 1, \label{E:rnkSb}\\
 \rank( \mathbf{S}_w ) &\le& N - C, \label{E:rnkSw}\\
 \rank( \mathbf{S}_t ) &\le& N - 1 \label{E:rnkSt},
\end{eqnarray}
for which, according to Lemma \ref{L:prodMatRnk}, the equality holds for SPD $\mathbf{K}$.
This is true, i.e. $\rank(\mathbf{K}) = \rank(\mathbf{\Phi}^T \mathbf{\Phi}) = N$,
for linearly independent training observation in the feature space,
e.g., when a strictly positive definite kernel function is used,
such as the Gaussian kernel \cite{HofSchSmo08}, or, when 
the linear kernel is used and the training observations are linearly independent
in the input space. In this case, the dimensionality $D$ of the discriminant subspace of AKDA is
$D  = C- 1$.  
Using the identity $\mathbf{R}_C^T \mathbf{R}_C = \mathbf{N}_C$, and
$\mathbf{C}_b \mathbf{J}_N = \mathbf{C}_w \mathbf{J}_N = \mathbf{0}_N$,
we can write
\begin{equation}
\mathbf{C}_{b} \mathbf{C}_{w} = \mathbf{0}_N, \;\; 
\mathbf{C}_{b} \mathbf{C}_{t} = \mathbf{C}_{b}, \;\; 
\mathbf{C}_{w} \mathbf{C}_{t} = \mathbf{C}_{w}. \nonumber 
\end{equation}
Therefore, $\mathbf{C}_{b}$ and $\mathbf{C}_{w}$ are orthogonal to each other.
Moreover, according to Lemma \ref{L:simultDiag} below, the NZEP of $\mathbf{C}_b$
provide the simultaneous reduction of $\mathbf{C}_b$, $\mathbf{C}_w$ and $\mathbf{C}_t$.
\begin{lemma} \label{L:simultDiag}
Let $\mathbf{A} $, $\mathbf{B} \in \mathbb{R}^{P \times P}$ be symmetric matrices, and 
$\mathbf{\Theta}  \in \mathbb{R}^{Q \times Q}$, $\mathbf{\Pi}  \in \mathbb{R}^{P \times Q}$,
$Q \leq P$, contain the NZEP of $\mathbf{A}$. If
$\mathbf{A} \mathbf{B} = \mathbf{A}$ 
then
$\mathbf{\Pi}^T \mathbf{B} \mathbf{\Pi} = \mathbf{I}_{Q}$. 
\end{lemma}
\textit{Proof}. The EVD of $\mathbf{A}$ can be used to write
$\mathbf{\Pi} \mathbf{\Theta} \mathbf{\Pi}^T \mathbf{B} = \mathbf{\Pi}
\mathbf{\Theta} \mathbf{\Pi}^T$.  
 Pre- and post-multiplying the above equation with
$\mathbf{\Theta}^{-1} \mathbf{\Pi}^T$ and $\mathbf{\Pi}$,
respectively, proofs the Lemma.

\subsection{Simultaneous reduction} \label{SS:simRdctAkda}

\begin{figure}[!tb]
\begin{center}
\begin{tabular}{c}
\includegraphics[width=.90\linewidth]{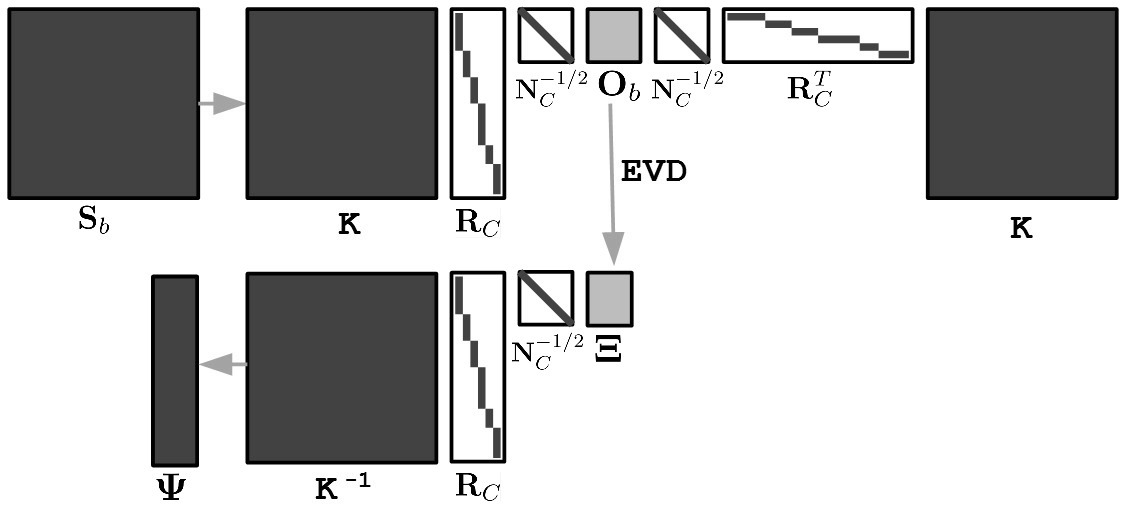}
\end{tabular}
\end{center}
\caption{Schematic representation of $\mathbf{S}_b$'s factorization and illustration of the AKDA algorithm (Algorithm \ref{ALG:AKDA}).} \label{fig:akdaVis}
\end{figure}

Taking into account (\ref{E:ObRnk}), (\ref{E:ObRng}), and the fact
that $\mathbf{O}_b$ (\ref{E:Ob}) is real symmetric idempotent, its EVD
can be expressed as
\begin{equation}
 \mathbf{\tilde{\Xi}}^T \mathbf{O}_{b} \mathbf{\tilde{\Xi}} =  \mathbf{\tilde{I}}_C, \nonumber
\end{equation}
where, $\mathbf{\tilde{\Xi}} = 
[ \mathbf{\Xi}, \frac{1}{N}\mathbf{\dot{n}}_C ] \in \mathbb{R}^{C \times C}$,
$\mathbf{\tilde{I}}_C = \diag(1, \dots, 1, 0) \in \mathbb{R}^{C \times C}$, and
$\mathbf{\tilde{\Xi}}^T \mathbf{\tilde{\Xi}} =  \mathbf{I}_C$.
Therefore, using the NZEP $(\mathbf{\Xi}, \mathbf{I}_{C-1})$
of $\mathbf{O}_{b}$ we can write
\begin{equation}
 \mathbf{\Xi}^T \mathbf{O}_{b} \mathbf{\Xi} =  \mathbf{I}_{C-1}. \label{E:ObEvdNzep}
\end{equation}
Almost always the number of classes $C$ is relatively small,
and thus, the computation of $\mathbf{\Xi}$ can be performed very
efficiently using various techniques.
Next, multiplying $\mathbf{C}_{b}$ (\ref{E:Cb}) from right and left
with $\mathbf{\Theta} \in \mathbb{R}^{N \times D}$,
\begin{equation}
 \mathbf{\Theta} = \mathbf{R}_C \mathbf{N}_C^{-1/2} \mathbf{\Xi}, \label{E:prjMatCb}
\end{equation}
and its transpose, and, using
(\ref{E:ObEvdNzep}) and the identity $\mathbf{R}_C^T \mathbf{R}_C = \mathbf{N}_C$
we get
\begin{equation}
 \mathbf{\Theta}^T \mathbf{C}_b \mathbf{\Theta} = \mathbf{\Xi}^T \mathbf{O}_{b} \mathbf{\Xi}
 \; = \; \mathbf{I}_{C-1}. \label{E:CbEvdNzep}
\end{equation}
Noting that $\mathbf{\Theta}^T \mathbf{\Theta} =  \mathbf{I}_{C-1}$, and taking into account
the idempotency of $\mathbf{C}_b$, and, (\ref{E:rnkCb}), (\ref{E:CbEvdNzep}),
we conclude that $(\mathbf{\Theta},  \mathbf{I}_{C-1})$ contain the NZEP of $\mathbf{C}_b$.
In the following we verify that the derived $\mathbf{\Theta}$ diagonalizes
the other two central factor matrices as well.
It is easy to show that $\mathbf{\Theta}$ spans the null space of $\mathbf{C}_w$
\begin{eqnarray}
 \mathbf{\Theta}^T \mathbf{C}_w \mathbf{\Theta} &=& \mathbf{\Theta}^T \mathbf{\Theta}
  - \mathbf{\Xi}^T \mathbf{N}_C^{-1/2} \mathbf{N}_C
\mathbf{N}_C^{-1} \mathbf{N}_C \mathbf{N}_C^{-1/2}  \mathbf{\Xi} \nonumber \\
   &=&  \mathbf{I}_{C-1} -  \mathbf{\Xi}^T  \mathbf{\Xi}  \; = \; \mathbf{0}_{C-1}. \label{E:CwEvdNzep}
\end{eqnarray}
Using the identity $\mathbf{R}_C^T \mathbf{1}_N = \mathbf{n}_C$
and the fact that $\mathbf{\tilde{\Xi}}$ is orthogonal, i.e.,
$\mathbf{\dot{n}}_C^T \mathbf{\Xi} = \left[ 0, \dots, 0 \right] \in \mathbb{R}^{1 \times C-1}$,
we can write
\begin{eqnarray}
 \mathbf{\Theta}^T \mathbf{C}_t \mathbf{\Theta} &=& \mathbf{\Theta}^T \mathbf{\Theta}
  - \mathbf{\Xi}^T \mathbf{N}_C^{-1/2} \mathbf{R}_C^T \mathbf{1}_N
   \mathbf{1}_N^T \mathbf{R}_C \mathbf{N}_C^{-1/2}  \mathbf{\Xi} \nonumber \\
   &=&  \mathbf{I}_{C-1} -  \mathbf{\Xi}^T  \mathbf{\dot{n}}_C \mathbf{\dot{n}}_C^T  \mathbf{\Xi}
\; = \; \mathbf{I}_{C-1}.  \label{E:CtEvdNzep}
\end{eqnarray}
The expansion coefficient matrix $\mathbf{\Psi}$ can then be computed by solving the following linear system
\begin{equation}
 \mathbf{K} \mathbf{\Psi} = \mathbf{\Theta}. \label{E:linSysPrjMatAkda}
\end{equation}
The linear system (\ref{E:linSysPrjMatAkda}) is consistent with a unique or multiple solutions when $\rank(\mathbf{K}) = N$
or $\rank([\mathbf{K}, \mathbf{\Theta}]) = \rank(\mathbf{K}) < N$, respectively.
As explained in the previous sections, $\mathbf{K}$ is SPD (and thus of full rank)
when the training observations are distinct and a strictly positive definite kernel is applied,
or when the linear kernel is used and the observations are linearly independent in the feature space.
Due to the fact that $\mathbf{K}$ is always SPD or SPSD, the Cholesky factorization
of $\mathbf{K}$ can be used to efficiently solve the above problem.
Specifically, two triangular linear systems are solved, $\mathbf{L} \mathbf{Y} = \mathbf{\Theta}$,
$\mathbf{L}^T \mathbf{\Psi} = \mathbf{Y}$ , where $ \mathbf{L}$ is the Cholesky factor of $\mathbf{K}$.
For ill-conditioned $\mathbf{K}$ a regularization step may be initially performed.
Using the dirived $\mathbf{\Psi}$ and (\ref{E:SigmbFac}), (\ref{E:SigmwFac}), (\ref{E:SigmtFac}),
(\ref{E:CbEvdNzep}), (\ref{E:CwEvdNzep}), (\ref{E:CtEvdNzep}), (\ref{E:linSysPrjMatAkda}),
it can be easily verified that $\mathbf{\Gamma}$ performs the 
simultaneous reduction of $\mathbf{\Sigma}_b$, $\mathbf{\Sigma}_w$ and $\mathbf{\Sigma}_t$
\begin{eqnarray}
  \mathbf{\Gamma}^T \mathbf{\Sigma}_b  \mathbf{\Gamma}   
  &=& \mathbf{\Psi}^T \mathbf{K} \mathbf{C}_b \mathbf{K} \mathbf{\Psi} 
  \; = \; \mathbf{\Theta}^T \mathbf{C}_b  \mathbf{\Theta}
  \; = \; \mathbf{I}_{C-1}, \label{E:akdaDimRedSb} \\
 \mathbf{\Gamma}^T \mathbf{\Sigma}_w  \mathbf{\Gamma}
  &=& \mathbf{\Psi}^T \mathbf{K} \mathbf{C}_w \mathbf{K} \mathbf{\Psi} 
  \; = \; \mathbf{\Theta}^T \mathbf{C}_w  \mathbf{\Theta}
  \; = \;  \mathbf{0}_{C-1}, \label{E:akdaDimRedSw} \\
  \mathbf{\Gamma}^T \mathbf{\Sigma}_t  \mathbf{\Gamma}
  &=& \mathbf{\Psi}^T \mathbf{K} \mathbf{C}_t \mathbf{K} \mathbf{\Psi} 
  \; = \;  \mathbf{\Theta}^T \mathbf{C}_t  \mathbf{\Theta} 
  \; = \; \mathbf{I}_{C-1}. \label{E:akdaDimRedSt}
\end{eqnarray}
We observe that AKDA is equivalent to KNDA \cite{Ye06P,Ye06J,Bodesheim13}.
However, requiring a cascade of SVD decompositions, KNDA is considerable less efficient than AKDA.
For SPD $\mathbf{K}$ the equality is true in the rank inequalities for 
$\mathbf{\Sigma}_b$ (\ref{E:rnkSb}), $\mathbf{\Sigma}_w$ (\ref{E:rnkSw}), $\mathbf{\Sigma}_t$ (\ref{E:rnkSt}),
and, thus, condition (\ref{C:rnkCndt}) is valid. In this case, AKDA shares the same properties with KUDA as well.
That is, the derived transformation matrix both whitens $\mathbf{\Sigma}_t$ and maximizes the between-class scatter in the null space of $\mathbf{\Sigma}_w$.
Further, by first performing the EVD of $\mathbf{\Psi}^T \mathbf{K} \mathbf{\Psi}
\rightarrow \mathbf{\tilde{\Pi}} \mathbf{\tilde{Q}} \mathbf{\tilde{\Pi}}^T$ and then setting $\mathbf{\Gamma}$
to $\mathbf{\Psi} \mathbf{\tilde{\Pi}} \mathbf{\tilde{Q}}^{-1/2}$, we get an equivalent transformation to KODA.
That is, $\mathbf{\Gamma}^T \mathbf{\Gamma} = \mathbf{I}$, and in this case, AKDA is equivalent to both KNDA and KODA.
\begin{algorithm}[!htb]
\caption{AKDA} \label{ALG:AKDA}
\begin{algorithmic}[1]
 {\REQUIRE $\mathbf{X}$, $Y_i, i=1,\dots,C$ (\ref{E:DataMatInpSpc}), $k(\cdot, \cdot)$ (\ref{E:KernelMap})}
 {\ENSURE $\mathbf{\Psi}$}
  \STATE Compute $\mathbf{O}_b$ (\ref{E:Ob}) and its eigenvector matrix $\mathbf{\Xi}$ (\ref{E:ObEvdNzep})
  \STATE Compute the eigenvector matrix $\mathbf{\Theta}$ of $\mathbf{C}_b$ using (\ref{E:prjMatCb})
  \STATE Compute the kernel matrix $\mathbf{K}$ using (\ref{E:KerMat})
  \STATE Solve (\ref{E:linSysPrjMatAkda}) to obtain $\mathbf{\Psi}$
\end{algorithmic}
\end{algorithm}

AKDA is summarized in Algorithm \ref{ALG:AKDA} and also illustrated in Figure \ref{fig:akdaVis}.
We observe that the proposed algorithm has very good numerical properties
as it consists of a few elementary matrix operations
(avoiding computing or operating directly to any of the associated scatter matrices),
and the symmetric QR and Cholesky factorization,
which are very stable algorithms for SPD or SPSD matrices.

\subsection{Binary classification} \label{SS:binaryAkda}

In the binary case, i.e. when $C=2$, most parts of AKDA (Algorithm \ref{ALG:AKDA}) can be computed analytically as described in the following.
Based on (\ref{E:ObRnk}), the rank of $\mathbf{O}_b$ (\ref{E:Ob}) equals to $C-1 = 1$.
Thus, taking additionally into account its idempotency, $\mathbf{O}_b$ has one positive eigenvalue $\lambda$, which equals to unity.
The corresponding eigenvector $\bg{\xi} = [\xi_1, \xi_2]^T$ is computed by solving the following homogeneous system
\begin{equation}
( \lambda \mathbf{I}_2 -  \mathbf{I}_2 + \mathbf{\dot{N}}_2 )\bg{\xi} = \mathbf{0}_2, \label{E:evecObHomEq}
\end{equation}
or equivalently $\mathbf{\dot{N}}_2  \bg{\xi} = \mathbf{0}_2$,
where
\begin{equation}
\mathbf{\dot{N}}_2 = \left[
 \begin{array}{cc} 
 N_1 & \sqrt{N_1 N_2} \\
 \sqrt{N_1 N_2} &  N_2 
\end{array} \right]. \nonumber
\end{equation}
We additionally require that the Euclidean norm of $\bg{\xi}$ equals to unity, i.e., $\xi_1^2 + \xi_2^2 = 1$.
From (\ref{E:evecObHomEq}) and the normality requirement above we arrive to
\begin{equation}
\xi_1 = \pm \sqrt{\frac{N_2}{N}}, \;\; \xi_2 = \mp \sqrt{\frac{N_1}{N}}. \label{E:evecCmpnCoreMatBin} 
\end{equation}
Next, using (\ref{E:prjMatCb}) the eigenvector $\bg{\theta}$ of $\mathbf{C}_b$ in the binary case is expressed as
\begin{equation}
  \bg{\theta} = { \small [\overbrace{ \pm \sqrt{\frac{N_2}{N_1 N}}, \dots, \pm \sqrt{\frac{N_2}{N_1 N}}}^{N_1}, 
\overbrace{ \mp \sqrt{\frac{N_1}{N_2 N}}, \dots, \mp \sqrt{\frac{N_1}{N_2 N}}}^{N_2} ]^T }, \label{E:evecCbBin}
\end{equation}
where we should note that its Euclidean norm equals to unity as well.
Similarly to (\ref{E:linSysPrjMatAkda}), the expansion coefficient vector $\bg{\psi}$ can then be computed by solving 
\begin{equation}
\mathbf{K} \bg{\psi} = \bg{\theta}. \label{E:linVecSysBin}
\end{equation}

\subsection{Computational complexity} \label{SS:akdaCmplx}

The computational complexity of AKDA in terms of flops \cite{Golub13} is analyzed in the following.
The calculation of $\mathbf{O}_b$ has complexity $O(C^2)$
based on the fact that the most intensive part during its computation
is the outer vector product $\mathbf{n}_C \mathbf{n}_C^T$.
The EVD of $\mathbf{O}_b$ using the symmetric QR algorithm has $9 C^3$ cost.
The computation of $\mathbf{\Theta}$ is $O(C)$ as it consists
of scaling the $i$th row of $\mathbf{\Xi}$
with $1/\sqrt{N_i}$ and then replicating it $N_i$ times.
The cost of the computation and the Cholesky factorization of $\mathbf{K}$
are $2 N^2 F$ and $N^3 / 3$, respectively, while the solution
of the two triangular systems has $2 N^2 (C-1)$ cost.
Summing up, the overall complexity of AKDA is
$N^3 / 3 + 2 N^2 (F + C-1) + 9 C^3 + O(C^2)+ O(C)$ or
$\frac{N^3}{3} + 2 N^2 (F + C-1)  + O(C^3)$. 
Note that in our implementation the computation of the scatter matrices is avoided.
For comparison, in the following we analyze the computational
complexity of conventional KDA \cite{Muller01,Mika00} and SRKDA \cite{Cai11_J},
which is one of its fastest variants.
The complexity of KDA is $(13 + 1/ 3) N^3 + 2 N^2 F$,
where, $2 N^3$, $N^3 / 3$, $2 N^3$, $9 N^3$ are the costs for
computing $\mathbf{S}_b$ and $\mathbf{S}_w$,
performing the Cholesky factorization of
$\mathbf{S}_w \rightarrow \mathbf{\tilde{L}} \mathbf{\tilde{L}}^{T}$, 
computing $\mathbf{\tilde{L}}^{-1} \mathbf{S}_b \mathbf{\tilde{L}}^{-T}$,
and performing the EVD of the above matrix
using the symmetric QR factorization, respectively.
Therefore, we observe that AKDA is approximately 40 times faster than KDA.
Concerning SRKDA, its complexity can be expressed as $N^3 / 3 + 2 N^2 F + 2 N^2 (C-1) + O(N^2) + N C^2 + C^3 / 3$
or $N^3 / 3 + 2 N^2 (F + C-1) + O(N^2)+ O(N)$, where,
$N C^2 + C^3 / 3$ is the cost of the Gram-Schmidt process applied to the matrix $\mathbf{\bar{C}}$, and,
$2 N^2 F$, $O(N^2)$, $N^3 / 3$, $2 N^2 (C-1)$ correspond to deriving $\mathbf{K}$, computing $\mathbf{\bar{K}}$ (\ref{E:GramCntrd}),
performing its Cholesky factorization, and solving $C-1$ linear equations, respectively.
Note that the computation of $\mathbf{\bar{K}}$ is $O(N^2)$ since all rows of $\mathbf{K} \mathbf{J}_N$
as well as all elements of $\mathbf{J}_N \mathbf{K} \mathbf{J}_N$ are identical.
Comparing the complexities of SRKDA and AKDA we observe
that they differ only in the last term of the expressions, set to $O(N^2) + O(N)$ and $O(C^3)$, respectively.
Therefore, it is expected that AKDA outperforms SRKDA in applications
with relatively large number of observations (i.e $N \gg C$) and where the term $O(N^2) + O(N)$
still plays a significant role in comparison to the $O(N^3)$ term in the overall complexity.

Turning back to the AKDA algorithm, we should note that its most intensive parts,
i.e., the calculation and Cholesky factorization of the kernel matrix,
can be both parallelized and performed at block level.
For instance, this is demonstrated in \cite{Chartampilas15,Chartampilas16},
where the use of a GPU accelerated tiled algorithm yielded approximately $O(N)$ complexity.

\section{AKSDA} \label{S:AKSDA}

Given $\mathbf{X}$ and subclass index sets $Y_{i,j}$ (\ref{E:DataMatInpSpc}),
AKSDA computes the transformation matrix $\mathbf{G} \in \mathbb{R}^{F \times \mathcal{D}}$
satisfying (\ref{E:ksdaTrCritFea}), (\ref{E:ksdaGepFea}).

\subsection{Factorization}

The between-subclass scatter matrix $\mathbf{\Sigma}_{bs}$ (\ref{E:SigmbsFea}) can be factorized as follows
\begin{equation}
\begin{array}{l}
 \displaystyle \mathbf{\Sigma}_{bs} =\sum_{i=1}^{C-1} \sum_{j=1}^{H_i}
\sum_{k=i+1}^{C} \frac{N_{k} N_{i,j}}{N}
\bg{\mu}_{i,j} \bg{\mu}_{i,j}^T\\
\displaystyle \quad \quad \quad + \sum_{i=1}^{C-1} \sum_{k=i+1}^{C} \sum_{l=1}^{H_k}
 \frac{N_{i} N_{k,l}}{N} \bg{\mu}_{k,l} \bg{\mu}_{k,l}^T\\
\displaystyle \quad \quad \quad - \sum_{i=1}^{C-1} \sum_{j=1}^{H_i}
\sum_{k=i+1}^{C} \sum_{l=1}^{H_k} \frac{N_{i,j} N_{k,l}}{N}
\bg{\mu}_{i,j} \bg{\mu}_{k,l}^T\\
\displaystyle \quad \quad \quad - \sum_{i=1}^{C-1} \sum_{j=1}^{H_i}
\sum_{k=i+1}^{C} \sum_{l=1}^{H_k} \frac{N_{i,j} N_{k,l}}{N}
\bg{\mu}_{k,l} \bg{\mu}_{i,j}^T\\
\displaystyle \quad \quad =  \sum_{i=1}^{C-1} \sum_{j=1}^{H_i}
\frac{(N - N_{i} ) N_{i,j}}{N} \bg{\mu}_{i,j} \bg{\mu}_{i,j}^T\\
\displaystyle \quad \quad \quad - \sum_{i=1}^{C-1} \sum_{j=1}^{H_i}
\sum_{k=1}^{C} \sum_{l=1}^{H_k} [\mathbf{E}]_{ij,kl} \frac{N_{i,j} N_{i,j}}{N}
\bg{\mu}_{k,l} \bg{\mu}_{i,j}^T\\
\displaystyle \quad \quad =   \mathbf{M}_H ( \mathbf{N}_H - \frac{1}{N} \mathbf{\grave{N}}_H \mathbf{N}_H
 - \frac{1}{N} \mathbf{\tilde{N}}_H  \circledast \mathbf{E} ) \mathbf{M}_H^T \\
 \displaystyle \quad \quad =   \mathbf{M}_H \mathbf{F} \mathbf{M}_H^T,
\end{array} \label{E:SigmbsFacPrelim}
\end{equation}
where, $\mathbf{M}_H$ is the matrix whose columns are the subclass means,
\begin{equation}
 \mathbf{M}_H =  [\bg{\mu}_{1,1}, \dots, \bg{\mu}_{C,H_C}] = \mathbf{\Phi} \mathbf{R}_H \mathbf{N}_H^{-1}, \label{E:Ms} \\
\end{equation}
$\mathbf{n}_H$, $\mathbf{N}_H$ are the so-called subclass strength vector and matrix, respectively,
\begin{eqnarray}
   \mathbf{n}_H &=& [  N_{1,1}, \dots, N_{C,H_C} ]^T, \label{E:subclassStrengthVec} \nonumber \\
 \mathbf{N}_H &=& \diag( N_{1,1}, \dots, N_{C,H_C} ), \label{E:subclassStrengthMat} \nonumber
\end{eqnarray}
$\mathbf{N}_H$, $\mathbf{\grave{N}}_H$ are the matrices resulting by the outer product of $\mathbf{n}_H$ with itself,
and the augmentation of the class strength matrix (\ref{E:classStrengthMat}) using the subclass information, respectively,
\begin{eqnarray}
  \mathbf{\tilde{N}}_H &=& \mathbf{n}_H \mathbf{n}_H^T, \nonumber \\
 \mathbf{\grave{N}}_H &=& \diag(\overbrace{N_1, \dots, N_1}^{H_1},  \dots, \overbrace{N_C, \dots, N_C}^{H_C} ), \nonumber
\end{eqnarray}
$\mathbf{R}_H \in \mathbb{R}^{N \times H}$ is the subclass indicator matrix whose element $[\mathbf{R}_H]_{n,ij}$
corresponding to the observation $\mathbf{x}_n$ and subclass $(i,j)$ is one if $n \in Y_{i,j}$ and zero otherwise,
$\mathbf{F}, \mathbf{E}$ are real symmetric ${H \times H}$ matrices whose elements $[\mathbf{F}]_{ij,kl}$,
$[\mathbf{E}]_{ij,kl}$, corresponding to subclasses $(i,j)$ and $(k,l)$ are defined as
\begin{eqnarray}
{[\mathbf{F}]}_{ij,kl}  &=& \frac{1}{N}\left \{
\begin{array}{ll} N_{i,j}(N -
N_i), & \mbox{if} \, (i,j) == (k,l) , \\
0 & \mbox{if} \, i = k, j \neq l , \\
- N_{i,j} N_{k,l} & \mbox{else},
\end{array} \right. \nonumber \\
 {[\mathbf{E}]}_{ij,kl}  &=& \left \{
 \begin{array}{ll} 
 0, & \mbox{if} \, i == k, \\
 1 & \mbox{else},
\end{array} \right. \nonumber
\end{eqnarray}
and $\circledast$ is the element-wise (or Hadamard) product of two equal-size matrices.
Similarly, $\mathbf{\Sigma}_{ws}$ (\ref{E:SigmwsFea}) can be factorized as
\begin{eqnarray}
\mathbf{\Sigma}_{ws} &=& \sum_{i=1}^{C} \sum_{j=1}^{H_i} \sum_{n \in
Y_{i,j}} \bg{\phi}_{n} \bg{\phi}_{n}^T - \sum_{i=1}^{C}
\sum_{j=1}^{H_i}
N_{i,j} \bg{\mu}_{i,j} \bg{\mu}_{i,j}^T \nonumber \\
 &=& \mathbf{\Phi} \mathbf{\Phi}^T -  \mathbf{M}_H \mathbf{N}_H
\mathbf{M}_H^T. \label{E:SigmwsFacPrelim}
\end{eqnarray}
Using (\ref{E:Ms}), $\mathbf{\Sigma}_{bs}$ (\ref{E:SigmbsFacPrelim}),
$\mathbf{\Sigma}_{ws}$ (\ref{E:SigmwsFacPrelim}), $\mathbf{S}_{bs}$
(\ref{E:SbsKer}) and $\mathbf{S}_{ws}$ (\ref{E:SwsKer}) can be expressed as
\begin{eqnarray}
 \mathbf{\Sigma}_{bs} &=& \mathbf{\Phi} \mathbf{C}_{bs} \mathbf{\Phi}^T,  \label{E:SigmbsFac} \\
 \mathbf{\Sigma}_{ws}  &=&  \mathbf{\Phi} \mathbf{C}_{ws} \mathbf{\Phi}^T.  \label{E:SigmwsFac} 
\end{eqnarray}
where
\begin{eqnarray}
\mathbf{C}_{bs} &=& \mathbf{R}_H \mathbf{N}_H^{-1/2} \mathbf{O}_{bs} \mathbf{N}_{H}^{-1/2}
\mathbf{R}_H^T, \label{E:Cbs} \nonumber \\
\mathbf{O}_{bs} &=& \mathbf{N}_H^{-1/2} \mathbf{F} \mathbf{N}_{H}^{-1/2}, \label{E:ObsPrelim} \\
\mathbf{C}_{ws}  &=& \mathbf{I}_N - \mathbf{R}_H \mathbf{N}_H^{-1} \mathbf{R}_H^T, \label{E:Cws} \nonumber
\end{eqnarray}
are the so-called between-subclass central and core matrices, and within-subclass matrix, respectively.
Element-wise, the core matrix can be expressed as
\begin{equation}
 {[\mathbf{O}_{bs}]}_{ij,kl}  = \frac{1}{N}\left \{
 \begin{array}{ll} N - N_i, & \mbox{if} \, (i,j) == (k,l) , \\
 0 & \mbox{if} \, i = k, j \neq l , \\
 - \sqrt{N_{i,j} N_{k,l}} & \mbox{else},
 \end{array} \right. \nonumber
\end{equation}
where ${[\mathbf{O}_{bs}]}_{ij,kl}$ is the element of $\mathbf{O}_{bs}$ corresponding to the subclasses $(i,j)$ and $(k,l)$.
Furthermore, using (\ref{E:SigmbsFac}), (\ref{E:SigmwsFac}), $\mathbf{S}_{bs}$
(\ref{E:SbsKer}) and $\mathbf{S}_{ws}$ (\ref{E:SwsKer}) can be written as
\begin{eqnarray}
 \mathbf{S}_{bs} &=& \mathbf{K} \mathbf{C}_{bs} \mathbf{K},  \label{E:SbsKerFac} \\
 \mathbf{S}_{ws}  &=&  \mathbf{K} \mathbf{C}_{ws} \mathbf{K}.  \label{E:SwsKerFac} 
\end{eqnarray}
Note that by evaluating further (\ref{E:ObsPrelim}), we can arrive to the following expression linking KSDA with kernel mixture discriminant analysis \cite{Gkalelis13NNLS} and KDA
\begin{equation}
\mathbf{O}_{bs}
      = \mathbf{I}_H - \frac{1}{N} \mathbf{\grave{N}}_H -  \mathbf{\dot{N}}_H  \circledast \mathbf{E}, \label{E:Obs}
\end{equation}
where, $\mathbf{\dot{N}}_H = \frac{  \mathbf{\dot{n}}_H \mathbf{\dot{n}}_H^T }{ \mathbf{\dot{n}}_H^T  \mathbf{\dot{n}}_H }$
and $\mathbf{\dot{n}}_H = [\sqrt{N_{1,1}}, \dots, \sqrt{N_{C,H_C}} ]^T$.
That is, alleviating the masking effect (i.e. using $\mathbf{E} = \mathbf{J}_H$) and
setting $\mathbf{\grave{N}}_H = \mathbf{0}_H$,
the matrices expressing the between-subclass variability (\ref{E:SigmbsFea}), (\ref{E:SbsKer})
become equivalent to their respective class counterparts (\ref{E:SigmbFea}), (\ref{E:SbKer}).

\subsection{Properties of the factorization}

It is trivial to show that the $\mathbf{C}_{bs}$, $\mathbf{C}_{ws}$,
$\mathbf{O}_{bs}$ and $\mathbf{F}$ are symmetric.
Considering each subclass as a graph vertex and noting that the sum
of each row/column of $\mathbf{F}$ is zero, $\mathbf{F}$ can be perceived as
the Laplacian of a graph of one connected component \cite{Bapat96}.
Thus, $\mathbf{F}$ is SPSD with
\begin{equation}
\rank( \mathbf{F} ) = H - 1, \;\; 
\rangesp( \mathbf{F} ) = \spanvs( \mathbf{1}_H )^\perp. \nonumber
\end{equation}
Using Lemma \ref{L:prodMatRnk}, the fact that $\mathbf{F}$ is SPSD, and,
$\mathbf{O}_{bs} \mathbf{\dot{n}}_H = \mathbf{N}_H^{-1/2} \mathbf{F} \mathbf{1}_H = \mathbf{0}$,
$\mathbf{x}^T \mathbf{O}_{bs} \mathbf{x}
=  \mathbf{y}^T \mathbf{F} \mathbf{y} \geq 0, \forall \mathbf{x} \neq \mathbf{0}$,
where $\mathbf{y} = \mathbf{N}_H^{-1/2} \mathbf{x}$, we conclude that $\mathbf{O}_{bs}$
is SPSD with
\begin{eqnarray}
\rank(  \mathbf{O}_{bs} ) &=& H - 1, \label{E:ObsRnk}\\
\rangesp( \mathbf{O}_{bs} ) &=& \spanvs(  \mathbf{\dot{n}}_H )^\perp.  \label{E:ObsRng}
\end{eqnarray}
Using Lemmas \ref{L:prodMatRnk}, \ref{L:sumMatRnk},
the idempotency of $\mathbf{R}_H \mathbf{N}_H^{-1} \mathbf{R}_H^T$,
and the fact that $\rank(\mathbf{R}_H \mathbf{N}_H^{-1} \mathbf{R}_H^T) = H$
we can show that
\begin{eqnarray}
\rank( \mathbf{S}_b ) &\le&\rank( \mathbf{C}_{bs}) =  H-1, \nonumber \\
\rank( \mathbf{S}_w ) &\le&\rank( \mathbf{C}_{ws}) = N - H. \nonumber
\end{eqnarray}
The equalities above hold for SPD $\mathbf{K}$, and in this case
$\mathcal{D} =  H - 1$.
Finally, using $\mathbf{R}_H^T \mathbf{R}_H = \mathbf{N}_C$, and
$\mathbf{C}_{bs} \mathbf{J}_N = \mathbf{C}_{ws} \mathbf{J}_N = \mathbf{0}_N$
we can write
\begin{equation}
\mathbf{C}_{bs} \mathbf{C}_{ws} = \mathbf{0}_N, \;\; 
\mathbf{C}_{bs} \mathbf{C}_{t} = \mathbf{C}_{b}, \;\; 
\mathbf{C}_{ws} \mathbf{C}_{t} = \mathbf{C}_{w}. \nonumber 
\end{equation}
Therefore, all three central factor matrices above can be simultaneously diagonalized.

\subsection{Simultaneous reduction}

Based on the fact that $\mathbf{O}_{bs}$ (\ref{E:Obs}) is real SPSD,
and using (\ref{E:ObsRnk}), (\ref{E:ObsRng}), its EVD can be expressed as
\begin{equation}
\mathbf{\tilde{U}}^T \mathbf{O}_{bs} \mathbf{\tilde{U}} =  \mathbf{\tilde{\Omega}}, \nonumber
\end{equation}
where, $\mathbf{\tilde{\Omega}}, \mathbf{\tilde{U}} \in \mathbb{R}^{H \times H}$, are the diagonal and orthogonal matrices containing the eigenpairs of $\mathbf{O}_{bs}$,
\begin{eqnarray}
 \mathbf{\tilde{\Omega}} &=& \left[ \begin{array}{rr}
	\mathbf{\Omega} & 0 \\
	0 & 0
\end{array} \right], \label{E:evlMatObs} \\
 \mathbf{\tilde{U}} &=&  \left[ \mathbf{U}, \textstyle \frac{1}{N} \mathbf{\dot{n}}_H \right]. \label{E:evdNzepObs}
\end{eqnarray}
Specifically, $\mathbf{\Omega}  \in \mathbb{R}^{H-1 \times H-1}$
contains the positive eigenvalues of $\mathbf{O}_{bs}$
in its diagonal, sorted in descending order, and the columns of $\mathbf{U} \in \mathbb{R}^{H \times H-1}$
are the corresponding eigenvectors, that is
\begin{equation}
 \mathbf{U}^T \mathbf{O}_{bs} \mathbf{U} = \mathbf{\Omega}. \label{E:ObsNzEvd}
\end{equation}
Due to the relatively small size of $\mathbf{O}_{bs}$ the computation of its NZEP $(\mathbf{U}, \mathbf{\Omega})$
can be performed efficiently using various techniques.
Let us define $\mathbf{V} \in \mathbb{R}^{N \times H-1}$ based on $\mathbf{U}$
as follows
\begin{equation}
 \mathbf{V} = \mathbf{R}_H \mathbf{N}_H^{-1/2} \mathbf{U}. \label{E:prjMatCbs}
\end{equation}
Noting that $\mathbf{V}$ has orthonormal columns and using (\ref{E:evdNzepObs})
and the identity
$\mathbf{R}_H^T \mathbf{R}_H =  \mathbf{N}_H$, we can easily show that
$(\mathbf{V}, \mathbf{\Omega})$ contain the NZEP of $\mathbf{C}_{bs}$,
and $\mathbf{V}$ spans the null space of $\mathbf{C}_{ws}$
\begin{eqnarray}
 \mathbf{V}^T \mathbf{C}_{bs} \mathbf{V} &=&  \mathbf{U}^T \mathbf{O}_{bs} \mathbf{U}
        \; = \; \mathbf{\Omega} \label{E:CbsSimRed}, \\
\mathbf{V}^T \mathbf{C}_{ws} \mathbf{V} &=& \mathbf{V}^T \mathbf{V} - \mathbf{V}^T
\mathbf{N}_H^{-1/2} \mathbf{N}_H \mathbf{N}_H^{-1} \mathbf{N}_H \mathbf{N}_H^{-1/2} \mathbf{V} \nonumber \\
    &=& \mathbf{0}_{H-1} \label{E:CwsSimRed}
\end{eqnarray}
Using the identity $\mathbf{R}_H^T \mathbf{J} \mathbf{R}_H =
\mathbf{n}_H \mathbf{n}_H^T$, and the orthogonality contition of $\mathbf{O}_{bs}$'s eigenvectors,
$\mathbf{\dot{n}}_H^T \mathbf{U} = \left[ 0, \dots, 0 \right] \in \mathbb{R}^{1 \times H-1}$,
it can be verified that $\mathbf{V}$ diagonalizes $\mathbf{C}_t$ as well
\begin{eqnarray}
 \mathbf{V}^T \mathbf{C}_t  \mathbf{V} &=&  \mathbf{V}^T \mathbf{V} - \mathbf{U}^T \mathbf{N}_H^{-1/2}
\mathbf{n}_H \mathbf{n}_H^T \mathbf{N}_H^{-1/2} \mathbf{U} \nonumber \\
  &=& \mathbf{V}^T \mathbf{V} - \mathbf{U}^T \mathbf{\dot{n}}_H \mathbf{\dot{n}}_H^T \mathbf{U}  
 \;= \; \mathbf{I}_{H-1}. \label{E:CtBsSimRed}
\end{eqnarray}
Based on $\mathbf{V}$, the expansion coefficient matrix $\mathbf{W}$ can then
be computed by solving the following linear system
\begin{equation}
 \mathbf{K} \mathbf{W} = \mathbf{V}. \label{E:linSysPrjMatAksda}
\end{equation}
The linear system above is consistent with a unique
or multiple solutions when $\rank(\mathbf{K}) = N$
or $\rank([\mathbf{K}, \mathbf{V}]) = \rank(\mathbf{K}) < N$, respectively.
As explained in Section \ref{SS:simRdctAkda}, $\mathbf{W}$
can be efficiently computed by performing the Cholesky factorization of $\mathbf{K}$
and then solving two triangular linear systems.
For ill-posed $\mathbf{K}$ a regularization step is first applied.
Using (\ref{E:SigmbsFac}), (\ref{E:SigmwsFac}), (\ref{E:SigmtFac}),
(\ref{E:CbsSimRed}), (\ref{E:CwsSimRed}), (\ref{E:CtBsSimRed}) and (\ref{E:linSysPrjMatAksda}),
it can be easily verified that the transformation matrix $\mathbf{G}$
provides the desired simultaneous reduction
\begin{eqnarray}
  \mathbf{G}^T \mathbf{\Sigma}_{bs}  \mathbf{G}
    &=&  \mathbf{V}^T \mathbf{C}_{bs} \mathbf{V} 
    =  \mathbf{\Omega}, \label{E:aksdaDimRedSbs} \\
 \mathbf{G}^T \mathbf{\Sigma}_{ws} \mathbf{G}
  &=& \mathbf{V}^T \mathbf{C}_{ws} \mathbf{V}
  \; = \; \mathbf{0}_{H-1}, \label{E:aksdaDimRedSws} \\
  \mathbf{G}^T \mathbf{\Sigma}_t  \mathbf{G}
  &=& \mathbf{V}^T \mathbf{C}_t  \mathbf{V}
  = \mathbf{I}_{H-1}. \label{E:aksdaDimRedSt}
\end{eqnarray}
That is, setting $\mathbf{\breve{\Lambda}} = \diag(+\infty, \dots, +\infty)$,
where division by zero is replaced by $+\infty$ as explained in Section \ref{S:Fundamentals},
or $\mathbf{\breve{\Lambda}} = \mathbf{\Omega}$ if $\mathbf{\Sigma}_t$
is used instead of $\mathbf{\Sigma}_{ws}$,
we can see that the computed $\mathbf{G}$
satisfies the GEP in (\ref{E:ksdaGepFea}),
and, thus, maximizes the AKSDA criterion in (\ref{E:ksdaTrCritFea}).
Note that here, in contrary to the simultaneous reductions
(\ref{E:akdaDimRedSb}), (\ref{E:akdaDimRedSw}), (\ref{E:akdaDimRedSt}) of AKDA,
the diagonal eigenvalue matrix $\mathbf{\Omega}$
does not necessarily equal to the identity matrix, and thus, the sum of the first
two reductions, (\ref{E:aksdaDimRedSbs}), (\ref{E:aksdaDimRedSws}),
does not equal the third (\ref{E:aksdaDimRedSt}).
Moreover, the fact that eigenvalues differ from each other widens the application domain of the method.
For instance, by retaining the eigenvectors corresponding to the 2 or 3 largest eigenvalues 
the proposed method can be used in data visualization tasks, offering an alternative perspective
in comparison to methods that use the directions that preserve most of the signal's variation \cite{Dhillon02}.
\begin{algorithm}[!htb]
\caption{AKSDA} \label{ALG:AKSDA}
\begin{algorithmic}[1]
 {\REQUIRE $\mathbf{X}$, $Y_{i,j}, i=1,\dots,C, j=1,\dots,H_i$ (\ref{E:DataMatInpSpc}), $k(\cdot, \cdot)$ (\ref{E:KernelMap})}
 {\ENSURE $\mathbf{W}$}
  \STATE Compute $\mathbf{O}_{bs}$ (\ref{E:Obs}) and its NZEP $(\mathbf{U}, \mathbf{\Omega})$ (\ref{E:ObsNzEvd})
  \STATE Compute the eigenvector matrix $\mathbf{V}$ of $\mathbf{C}_{bs}$ using (\ref{E:prjMatCbs})
  \STATE Compute the kernel matrix $\mathbf{K}$ using (\ref{E:KerMat})
  \STATE Solve (\ref{E:linSysPrjMatAksda}) to obtain $\mathbf{W}$
\end{algorithmic}
\end{algorithm}

AKSDA is summarized in Algorithm \ref{ALG:AKSDA}.
We can see that the computation of $\mathbf{W}$
consists of a few elementary matrix operations,
the EVD of a relatively small-sized matrix and the symmetric Cholesky factorization.
Thus, similar to AKDA (Algorithm \ref{ALG:AKDA}),
AKSDA exhibits very good numerical stability and computational efficiency.

\subsection{Computational complexity} \label{SS:aksdaCmplx}

The computational complexity of AKSDA is
$\frac{N^3}{3} + 2 N^2 (F + H-1)  + O(N) + O(H^3)$, 
which consists of the cost applying k-means \cite{Telgarsky10}
to create a subclass division (the $O(N)$ term above),
and similar to AKDA, 
the costs for computing $\mathbf{K}$ and its Cholesky factorization, 
computing $\mathbf{O}_{bs}$ and its EVD,
formulating $\mathbf{V}$, and solving the two triangular systems.
On the other hand, as in KDA (Section \ref{SS:akdaCmplx}),
the complexity of KSDA is $\frac{40}{3} N^3 + 2 N^2 F + O(N^2)$, where the last term 
is the cost for applying the nearest-neighbor-based partitioning procedure of KSDA \cite{Zhu06,You11}.
Therefore, for large-scale datasets we can see that AKSDA is approximately 40 times faster than KSDA, while 
for average-sized datasets, where the $O(N^2)$ term still plays a significant role, AKSDA provides a further speedup. 
We should note that similarly to AKDA, AKSDA can be fully parallelized,
offering an approximately $O(N)$ complexity \cite{Chartampilas15,Chartampilas16}.

\section{Experiments} \label{S:exprms}

In this section the experimental analysis of the proposed methods is performed.
Specifically, in Section \ref{S:toyEx}, we provide a toy example in order to demonstrate the effectiveness
and gain insight info of the proposed acceleration framework.
Then, in Section \ref{SS:expEval}, we proceed to the experimental evaluation of the proposed AKDA and AKSDA
in various image and video datasets for the tasks of event, concept and object detection.

\subsection{Datasets} \label{SS:datasets}

For the experimental analysis we utilize two TRECVID MED datasets \cite{trecvidawad2016,habibian13,Chartampilas15}
and the cross-dataset collection \cite{Tommasi14}, as described in the following.

\subsubsection{TRECVID MED} \label{SS:trecvidMed}

For event detection in video, the following TRECVID MED datasets \cite{trecvidawad2016} are used:
a) \textit{med10}: This is the MED 2010 dataset consisting of 1745 training and 1742 testing videos belonging to one of 3 target events
or to the ``rest-of-world" event category. 
Improved dense trajectories features \cite{Wang13,Chartampilas15} are extracted to represent each video in the input space $\mathbb{R}^L$, where $L= 101376$.
b) \textit{med--hbb}: This dataset has been created using a subset of the MED 2012 video corpus and the respective publicly available partitioning provided in \cite{habibian13}.
It consists of 25 target events and 13274 videos, divided to 8824 videos for training and 4425 for testing. 
Similarly to med10, each video is represented in $\mathbb{R}^{101376}$ using the improved dense trajectories features.

\subsubsection{Cross-dataset collection} \label{SS:medHbbDb}

For the tasks of object and concept detection,
11 publicly available datasets from the cross-dataset collection \cite{Tommasi14} are used,
briefly described in the following:
a) \textit{AwA}: This dataset provides over 30000 images belonging to one of 50 classes 
matching the 50 animal categories in Osherson’s animal/attribute matrix.
b) \textit{Ayahoo}: It consists of 12 classes and 2237 object images collected using the Yahoo! image search. 
c) \textit{Bing}: Augments the classes of the Caltech256 
dataset with 300 weakly-labeled Internet images collected using Bing.
d) \textit{Caltech101}: It contains object images belonging 
to 101 categories, such as airplane, chair and crocodile.
e) \textit{Caltech256}: It contains 257 object classes, offering a wider variety of objects in comparison to Caltech101. 
f) \textit{Eth80}: It consists of 8 object classes, where objects are recorded from 41 different views. 
g) \textit{Imagenet}: This is another large-scale dataset which aims to populate the majority 
of the 80,000 synsets of WordNet with an average of 500-1000 full resolution images.
h) \textit{Msrcorid}: It is the Microsoft research Cambridge object recognition image database 
consisting of 22 classes and 36 to 652 images per class.
i) \textit{Office}: It contains images of various resolutions from Amazon.com and office environment captured using a webcam and 
digital LSR camera.
j) \textit{Pascal07}: It is the benchmark dataset of the 2007 visual object classes (VOC) challenge, 
containing consumer photographs from the flickr2 photo-sharing web-site.
k) \textit{Rgbd}: It is a large-scale, hierarchical multi-view object dataset collected using an RGB-D camera. 
\begin{table}[!htb]
\caption{Datasets in the cross-dataset collection}
\begin{center}
{\footnotesize
\begin{tabular}{lcccccc}
\toprule
     & & \multicolumn{2}{c}{\textit{10Ex}} && \multicolumn{2}{c}{\textit{100Ex}} \\
\cmidrule{3-4} \cmidrule{6-7}
     & \textit{\# classes} & \textit{\# train} & \textit{\#  test} && \textit{\# train} & \textit{\# test}\\
\midrule
  \textit{AwA}            & 50 & 500 & 30233 && 4941 & 25792 \\
  \textit{ayahoo}       & 12 & 120 & 2117 && 988 & 1249 \\
  \textit{bing}            & 257 & 2570  & 118352 && 25698 & 95224 \\
  \textit{caltech101} & 101 & 1010  & 7721 && 3539 & 5192 \\
  \textit{caltech256} & 257 & 2570 & 28699 && 14106 & 17163\\
  \textit{eth80}         & 80 & 800 & 2480 && 1680 & 1600 \\
  \textit{imagenet}   & 118 & 1180 & 164418 && 11762 & 153836 \\
  \textit{mscorid}      & 22 & 220 & 3912 && 1497 & 2635 \\
  \textit{office}         & 91 & 836 & 3278 && 2075 & 2039 \\
  \textit{pascal07}     & 20 & 200 & 14119 && 1997 & 12322 \\
  \textit{rgdb}            &  51 & 510 & 207410 && 5100 & 202820 \\
\bottomrule
\end{tabular}}
\end{center}
\label{tbl:crossDtstDistrib}
\end{table}
The cross-dataset collection has been created using the original image datasets and various feature extraction procedures.
Here, the image descriptors corresponding to the 4096 neurons of the 6-th layer
of DeCAF \cite{Donahue14} deep learning implementation provided in \cite{Tommasi14} are used.

For the evaluation experiments in Section \ref{SS:expEval},
two experimental conditions and the respective training/testing set divisions are designed for each dataset.
Specifically, we vary the number of training observation per class, following an evaluation protocol similar to the
10Ex and 100Ex query conditions in the TRECVID MED evaluation challenge \cite{trecvidawad2016}:
\begin{itemize}
 \item \textit{10Ex}: 10 positive observations are randomly selected from each class to create the training set,
and the rest of the observations are used for testing.
 \item \textit{100Ex}: Similar to 10Ex, but now 100 positive observations per class are randomly selected to formulate the training set.
\end{itemize}
For classes where the total number of observations is less than 20 or 200,
for the 10Ex or 100Ex condition, respectively,
half of the observations are used for training, and the rest for testing.
The number of classes and the size of training and testing sets for each dataset and experimental condition are shown in Table \ref{tbl:crossDtstDistrib}.

\subsection{Toy example} \label{S:toyEx}

The purpose of this example is to demonstrate the effectiveness and gain insight of the proposed AKDA.
In the following, we show the different steps of AKDA in order to identify a discriminant subspace for the rgdb class apple, called hereafter the target class.
We utilize the rgbd training set under the 100Ex experimental condition, consisting of 
$N_1 = 100$ and $N_2 = 5000$ observations belonging to the target and rest-of-world class, respectively, as explained in Section \ref{SS:datasets}.
That is, $N = 5100$ observations are provided in total.
\begin{figure}[!htb]
\begin{center}
\begin{tabular}{c}
\includegraphics[width=.90\linewidth]{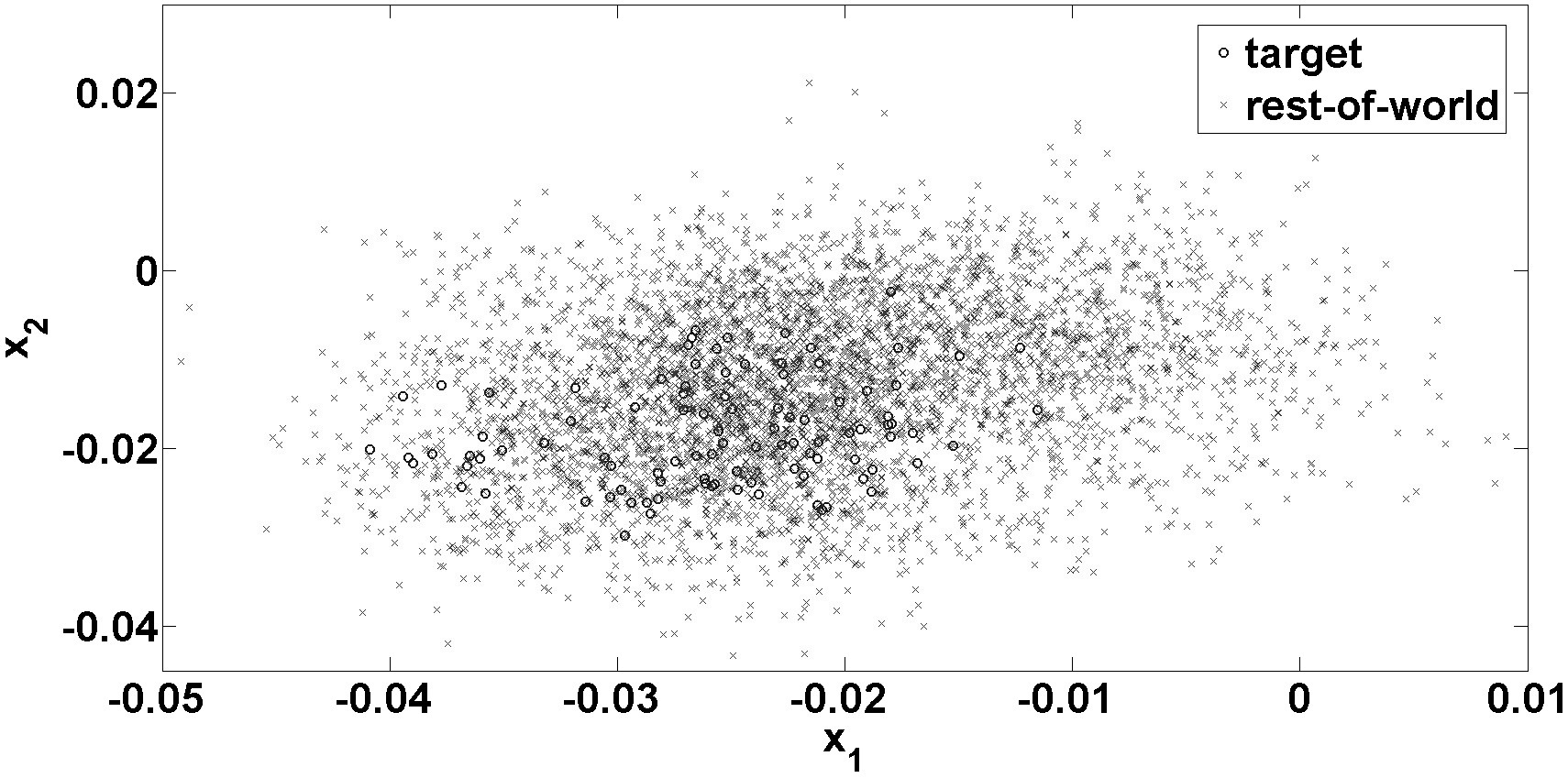}
\end{tabular}
\end{center}
\caption{Scatter plot created using the first and second components of the rgbd training observations in the input space $\mathbb{R}^{4096}$.} \label{fig:toyScatterEx}
\end{figure}
In Figure \ref{fig:toyScatterEx}, the scatter plot of the first and second dimension of the input space $\mathbb{R}^{L}$, $L=4096$,
along all training observations is depicted.
From this plot we observe that the class distributions overlap in the specified subspace. 
The same conclusion can be drawn by inspecting similar 2D scatter plots for the rest of the input space dimensions.

We now proceed to the application of AKDA in the above problem.
The eigenvector $\bg{\xi}$ (\ref{E:evecCmpnCoreMatBin}) and $\bg{\theta}$ (\ref{E:evecCbBin}) corresponding to the nonzero eigenvalues
of the core matrix $\mathbf{O}_b$ and the between-class central factor matrix $\mathbf{C}_b$, respectively,
are
\begin{eqnarray}
  \bg{\xi} &=& [-0.9901, 0.1400]^T, \nonumber \\
  \bg{\theta} &=& { \small [\overbrace{ -0.09901, \dots, -0.09901}^{100}, 
\overbrace{ 0.00198, \dots, 0.00198}^{5000} ]^T }. \label{E:evecCbToy} \nonumber
\end{eqnarray}
Subsequently, the kernel matrix $\mathbf{K}$ (\ref{E:KerMat}) and the projection vector $\bg{\psi}$ (\ref{E:linVecSysBin}) are derived,
where for $\mathbf{K}$ the linear kernel is used.
The overall learning time for AKDA is approximately 2.25 seconds, which is dominated by the 
time to compute the kernel matrix (1.62 secs) and the time to solve the linear system (0.63 secs).
This is a significant speedup over KDA, considering that its learning time in this task is 140.96 seconds.

The training observations $\mathbf{x}_n \in \mathbb{R}^L$ are then projected into the 1D subspace using
$z_n = \bg{\gamma}^T \mathbf{x}_n$, $n=1,\dots, N$, where $\bg{\gamma} = \mathbf{X} \bg{\psi}$,
and $\mathbf{X} \in \mathbb{R}^{L \times N}$ contains the observations of the training set.
\begin{figure}[!htb]
\begin{center}
\begin{tabular}{c}
\includegraphics[width=.90\linewidth]{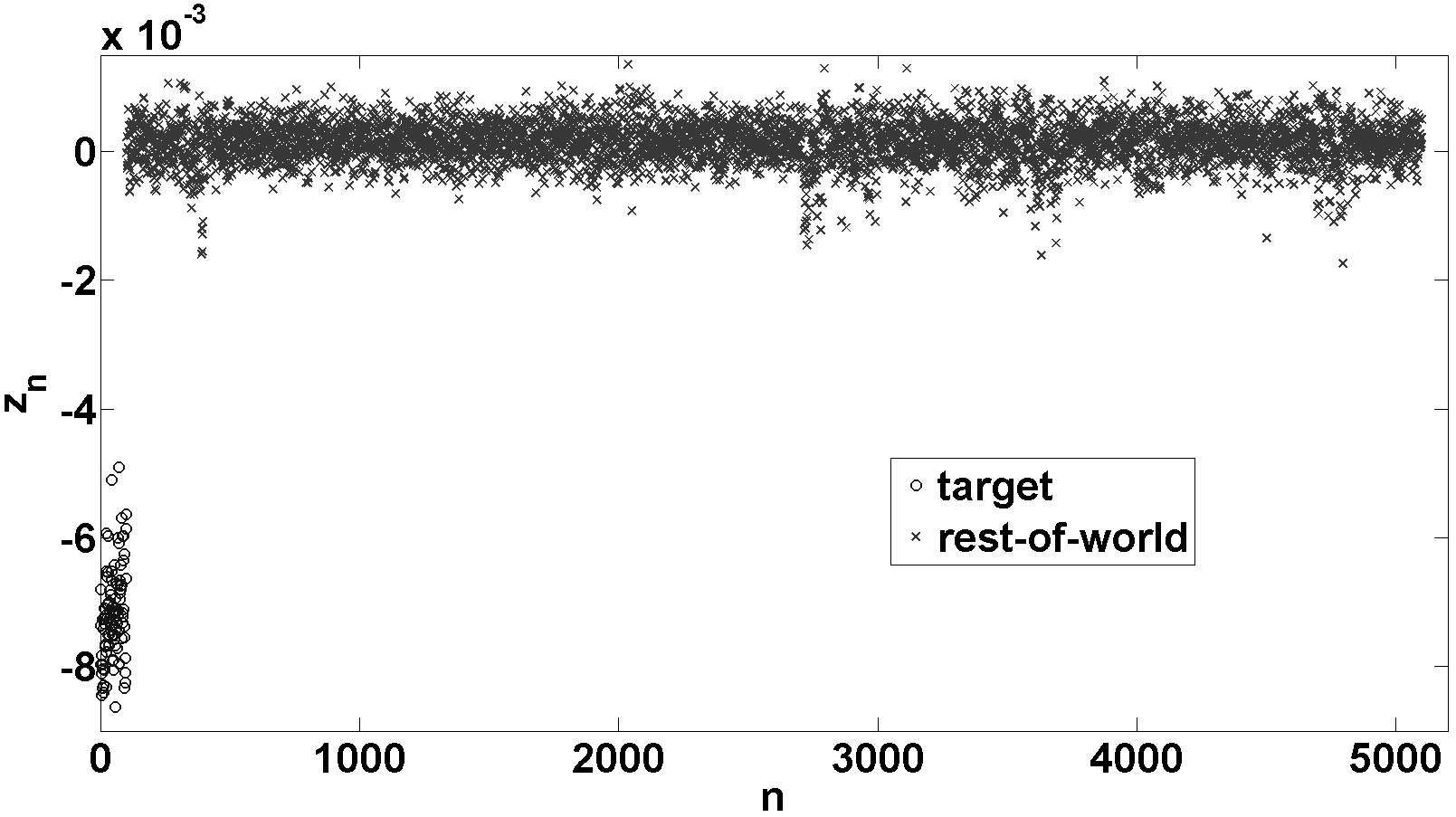}
\end{tabular}
\end{center}
\caption{Training observations of rgbd dataset projected in the 1D subspace identified using AKDA.} \label{fig:toyAkdaSubspace}
\end{figure}
Figure \ref{fig:toyAkdaSubspace} depicts the values $z_n$ derived from the application of AKDA in the rgdb training set.
We observe that in the resulting subspace the two classes are well separated.
Due to this fact, we conclude that the combination of AKDA with linear classifiers such as LSVM can provide an effective classification framework.
The main advantages of such an approach are the overall acceleration provided by the use of AKDA
and the ability to alleviate the curse of dimensionality problem of traditional classifiers,
as shown by the evaluation results in Section \ref{SS:results}.

\subsection{Experimental evaluation} \label{SS:expEval}

In this section, the proposed AKDA and AKSDA are compared with 7 popular discriminant analysis approaches,
specifically, 2 linear (PCA, LDA), 3 kernel (KDA, GDA, SRKDA), and 2 kernel subclass-based (KSDA, GSDA).
In order to permit their comparison, the DR methods are further combined with a binary linear support vector machine (LSVM).
That is, one LSVM is trained for each class in the discriminant subspace derived with the respective DR method
and subsequently used to assign a confidence score to each test observation.
The above methods are also compared with LSVM and KSVM applied directly in the input space.
The evaluation is performed on the datasets described in Section \ref{SS:datasets}.

\subsubsection{Setup} \label{SS:ExpSetup}

The evaluation in the 2 TRECVID MED datasets is performed by directly utilizing the video annotations
and training/testing set divisions provided in \cite{habibian13,Chartampilas15,trecvidawad2016}. 
For the cross-datasets collection the two experimental conditions
and the respective training/testing set divisions described in Table \ref{tbl:crossDtstDistrib} are used.

For GSDA and KSDA the Matlab implementations provided in \cite{Gkalelis14} and \cite{You11} are utilized, respectively,
while the rest of the methods are coded in unoptimized Matlab.
For LSVM and KSVM we use the C++ libraries provided in \cite{libsvm}.
Kernel methods utilize the Gaussian radial basis function
$k(\mathbf{x}_n, \mathbf{x}_\nu) = \exp(- \varrho \| \mathbf{x}_n - \mathbf{x}_\nu \|^2), \varrho \in \mathbb{R}_+$
as base kernel.
For dividing classes to subclasses, KSDA exploits the nearest neighbor-based clustering approach described in \cite{Zhu06},
while for AKSDA and GSDA the k-means clustering procedure presented in \cite{Gkalelis14} is used.
The different approaches are optimized using 3-fold cross-validation (CV),
where at each fold the training set is randomly split to 30\% learning set and 70\% validation set.
Specifically, the kernel parameter $\varrho$, the SVM penalty term $\varsigma$, and the total number of subclasses $H$ are searched in $\{0.01, 0.1, 0.6\} \cup  \{1, 1.5, \ldots , 7\}$, $\{0.1, 1, 10, 100\}$, and $\{2, 3, \dots, 5\}$, respectively.
To alleviate the singularity of the centered kernel matrix in GDA, SRKDA and GSDA,
regularization is applied, where the regularization constant $\epsilon$ is set to $10^{-3}$.
The same constant is also used for the regularization of ill-posed kernel (AKDA, AKSDA), within-class
or total scatter matrix (PCA, LDA, KDA, KSDA).

Each method is evaluated for each dataset and experimental condition.
The performance of the $m$th method in a specified experiment is measured using the mean average precision (MAP) $\varpi_m$, and 
training and testing time speedup over KDA, $\tilde{\vartheta}_m$, $\tilde{\varphi}_m$, respectively,
defined as 
\begin{equation}
\varpi_m = \frac{1}{C} \sum_{i=1}^C \varpi_{m,i}, \; \; 
\tilde{\vartheta}_m = \frac{\vartheta_{KDA}}{\vartheta_m}, \; \; 
\tilde{\varphi}_m = \frac{\varphi_{KDA}}{\varphi_m},   \nonumber
\end{equation}
where, $\vartheta_m = \sum_{i=1}^C \vartheta_{m,i}$, $\varphi_m = \sum_{i=1}^C \varphi_{m,i}$,
and,  $\varpi_{m,i}$, $\vartheta_{m,i}$, $\varphi_{m,i}$, are the average precision, 
training time and testing time of the $m$th method with respect to the $i$th class, respectively.
For the training time $\vartheta_{m,i}$, we consider only the time for building the $i$th classifier
with the parameters fixed to the optimal parameters derived
using the CV procedure, i.e., the CV time is excluded.
The evaluation is performed using Intel i7 3770K@3.5Ghz CPU, 32 GB RAM workstations with 64-bit Windows 7.

\subsubsection{Results} \label{SS:results}

\begin{table*}[!htb]
\caption{MAP rates on TRECVID MED datasets}
\begin{center}
{\footnotesize
\begin{tabular}{lccccccccccccc}
\toprule
    & \multicolumn{3}{c}{Linear methods} && \multicolumn{5}{c}{Kernel methods}
    && \multicolumn{3}{c}{Kernel subclass methods} \\
    \cmidrule{2-4} \cmidrule{6-10} \cmidrule{12-14}
     & \textit{PCA} & \textit{LDA} & \textit{LSVM} && \textit{KDA} & \textit{GDA} 
     & \textit{SRKDA} & \textit{AKDA} & \textit{KSVM} && \textit{KSDA} & \textit{GSDA}
     & \textit{AKSDA}\\
\midrule
  \textit{med10}         & 55.09\%  & 56.42\% & 54.53\% && 54.36\% & 56.77\% & 55.95\% & \textbf{57.64}\% & 56.32\% && 55.98\% & 55.47\%  &  \textbf{57.64}\% \\ 
  \textit{med--hbb}       &  34.44\% & 38.15\% & 39.99\% && 41.61\% & 41.23\% & 41.22\% & 42.55\% & 40.39\% && 43.15\% & 44.73\% &  \textbf{45.51\%} \\
\bottomrule
\end{tabular}}
\end{center}
\label{tbl:trecMedMap}
\end{table*}
\begin{table*}[!htb]
\caption{MAP rates on cross-dataset collection -- 10Ex condition}
\begin{center}
{\footnotesize
\begin{tabular}{lccccccccccccc}
\toprule
    & \multicolumn{3}{c}{Linear methods} && \multicolumn{5}{c}{Kernel methods}
    && \multicolumn{3}{c}{Kernel subclass methods} \\
    \cmidrule{2-4} \cmidrule{6-10} \cmidrule{12-14}
     & \textit{PCA} & \textit{LDA} & \textit{LSVM} && \textit{KDA} & \textit{GDA	} 
     & \textit{SRKDA} & \textit{AKDA} & \textit{KSVM} && \textit{KSDA} & \textit{GSDA}
     & \textit{AKSDA}\\
\midrule
  \textit{AwA}            & 39.64\% & 25.79\% & 39.43\% && 41.78\% & 43.04\% & 43.13\% & 44.68\% & 43.03\%&& 42.13\% & 44.06\%  &  \textbf{44.89\%} \\ 
  \textit{ayahoo}       & 74.33\% & 70.38\% & 73.36\% && 72.77\% & 75.22\% & 75.62\% & 75.50\% & 74.54\% && 75.82\% & 73.88\% &  \textbf{76.43\%} \\
  \textit{bing}            & 7.11\%   & 2.12\%   &  9.90\%  && 11.99\%  & 12.33\% & 12.34\% & \textbf{12.44\%} & 11.29\% && 12.33\% & 12.41\% & \textbf{12.44\%} \\
  \textit{caltech101} & 76.41\% & 70.06\% & 77.81\% && 77.26\% & 77.88\% & 80.09\% & 80.02\% & 78.15\% && 79.32\% & 80.25\% & \textbf{80.48\%} \\
  \textit{caltech256} & 40.32\% & 17.73\% & 50.04\% && 51.09\% & 52.45\% & 52.11\% & 53.61\% & 51.81\% && 53.32\% & 53.34\% & \textbf{53.67\%} \\
  \textit{eth80}         & 76.90\% & 76.43\% & 81.13\% && 78.90\% & 79.11\% & 81.41\% & 81.88\% & 79.21\% && 81.36\% & 79.63\% & \textbf{82.01\%} \\
  \textit{imagenet}   & 35.24\% & 20.55\% & 39.91\% && 43.08\% & 43.11\% & 44.14\% & 43.57\% & 42.16\% && 42.70\% & 42.83\% & \textbf{43.62\%} \\
  \textit{mscorid}      & 87.52\% & 86.83\% & 87.19\% && 86.11\% & 88.14\% & 88.66\% & 88.66\% & 88.01\% && 87.97\% & 88.43\% & \textbf{89.21\%} \\
  \textit{office}         & 65.18\% & 53.07\% & 66.04\% && 61.36\% & 63.24\% & 63.99\% & 65.27\% & 65.31\% && 63.76\% & 65.80\% & \textbf{68.37\%} \\
  \textit{pascal07}    & 29.40\% & 25.19\% & 30.24\% && 33.75\% & 32.53\% & 31.16\% & 33.77\% & 32.17\% &&  33.07\% & 32.95\% & \textbf{34.22\%} \\
  \textit{rgdb}           & 91.33\% & 91.57\% & 89.03\% && 89.51\% & 84.70\% & 91.80\% & 94.77\% & 94.81\% && 93.26\% & 92.09\% & \textbf{95.17\%} \\
\midrule
  \textit{Average}     & 56.67\% & 49.07\% & 58.46\% && 58.87\% & 59.24\% & 60.41\% & 61.29\% & 60.05\% && 60.46\% & 60.52\% & \textbf{61.86\%} \\
\bottomrule
\end{tabular}}
\end{center}
\label{tbl:crossDtStMap10Ex}
\end{table*}
\begin{table*}[!htb]
\caption{MAP rates on cross-dataset collection -- 100Ex condition}
\begin{center}
{\footnotesize
\begin{tabular}{lccccccccccccc}
\toprule
    & \multicolumn{3}{c}{Linear methods} && \multicolumn{5}{c}{Kernel methods}
    && \multicolumn{3}{c}{Kernel subclass methods} \\
    \cmidrule{2-4} \cmidrule{6-10} \cmidrule{12-14}
     & \textit{PCA} & \textit{LDA} & \textit{LSVM} && \textit{KDA} & \textit{GDA} 
     & \textit{SRKDA} & \textit{AKDA} & \textit{KSVM} && \textit{KSDA} & \textit{GSDA}
     & \textit{AKSDA}\\
\midrule
  \textit{AwA}            & 43.61\% & 23.37\% & 58.29\% && 61.96\% & 62.06\% & 62.97\% & 67.41\% & 61.35\% && 62.97\% & 64.15\% & \textbf{67.43\%} \\
  \textit{ayahoo}       & 77.04\% & 74.60\% & 77.04\% && 86.44\% & 87.83\% & 87.18\% & 88.01\% & 85.94\% && 87.88\% &87.39\% & \textbf{88.30\%} \\
  \textit{bing}            & 10.76\% & 15.21\% & 17.16\%  && 25.74\% & 26.31\% & 26.36\% & 27.74\% & 26.04\% && 26.15\% & 26.24\% & \textbf{27.82\%} \\
  \textit{caltech101} & 77.00\% & 77.51\% & 82.45\% && 82.08\% & 83.70\% & 83.24\% & 84.34\% & 83.32\% && 84.48\% & 84.74\% & \textbf{85.65\%} \\
  \textit{caltech256} & 41.65\% & 49.20\% & 63.59\% && 65.03\% & 69.35\% & 70.54\% & 70.38\% & 67.05\% && 69.02\% & 70.93\% & \textbf{71.99\%} \\
  \textit{eth80}         & 76.91\% & 76.77\% & 82.06\% && 83.28\% & 80.53\% & 81.11\% & 84.58\% & 82.27\% && 85.57\% & 81.44\% & \textbf{85.15\%} \\
  \textit{imagenet}   & 38.87\% & 41.80\% & 52.59\% && 63.22\% & 63.24\% & 63.94\% & 65.70\% & 63.66\% && 63.79\% & 65.50\% & \textbf{66.46\%} \\
  \textit{mscorid}      & 90.98\% & 87.10\% & 92.94\% && 93.16\% & 89.10\% & 93.11\% & 95.43\% & 92.92\% && 92.47\% & 93.87\% & \textbf{95.43\%} \\
  \textit{office}         & 66.94\% & 54.97\% & 70.40\% && 67.71\% & 67.44\% & 69.68\% & 71.70\% & 69.81\% && 74.20\% & 73.63\% & \textbf{75.18\%} \\
  \textit{pascal07}    & 29.40\% & 25.19\% & 35.80\% && 35.75\% & 35.30\% & 34.55\% & 36.19\% & 35.30\% && 36.07\% & 37.87\% & \textbf{39.50\%} \\
  \textit{rgdb}           & 83.02\% & 96.50\% & 99.49\% && 97.69\% & \textbf{99.96\%} & 99.92\% & \textbf{99.97\%} &  99.83\% &&  97.67\% & 99.89\% & 99.95\% \\
\midrule
  \textit{Average}     & 57.83\% & 56.57\% & 66.53\% && 69.28\% & 69.53\% & 70.23\% & 71.95\% & 69.81\% && 70.93\% & 71.42\% & \textbf{72.99}\% \\
\bottomrule
\end{tabular}}
\end{center}
\label{tbl:crossDtStMap100Ex}
\end{table*}
\begin{table*}[!htb]
\caption{Training/testing time speedup on TRECVID MED datasets}
\begin{center}
{\footnotesize
\begin{tabular}{lccccccccccccc}
\toprule
     & \multicolumn{3}{c}{Linear methods} && \multicolumn{5}{c}{Kernel methods}
    && \multicolumn{3}{c}{Kernel subclass methods} \\
    \cmidrule{2-4} \cmidrule{6-10} \cmidrule{12-14}
     & \textit{PCA} & \textit{LDA} & \textit{LSVM} && \textit{KDA} & \textit{GDA} 
     & \textit{SRKDA} & \textit{AKDA} & \textit{KSVM} && \textit{KSDA} & \textit{GSDA}
     & \textit{AKSDA}\\
\midrule
  \textit{med10}            & 0.4/1 & 0.9/1.2 & 0.03/0.4 && 1/1 &  0.4/1 &  1/1 & \textbf{2}/\textbf{1.3} &  0.03/0.4 &&  0.002/1 &  0.4/1  &  \textbf{2}/1  \\
  \textit{med--hbb}       & 0.7/1.5 & 1.3/\textbf{2.9} &  0.1/0.2 &&  1/1 &  0.1/0.1  &  3.6/0.1 & \textbf{7.1}/1.2 &  0.06/0.1  &&  0.01/0.8 &  0.1/0.1  &  6.9/0.9 \\
\bottomrule
\end{tabular}}
\end{center}
\label{tbl:trecMedTrnTstTimes}
\end{table*}
\begin{table*}[!htb]
\caption{Training/testing time speedup on cross-dataset collection -- 10Ex condition}
\begin{center}
{\footnotesize
\begin{tabular}{lccccccccccccc}
\toprule
     & \multicolumn{3}{c}{Linear methods} && \multicolumn{5}{c}{Kernel methods}
    && \multicolumn{3}{c}{Kernel subclass methods} \\
    \cmidrule{2-4} \cmidrule{6-10} \cmidrule{12-14}
     & \textit{PCA} & \textit{LDA} & \textit{LSVM} && \textit{KDA} & \textit{GDA} 
     & \textit{SRKDA} & \textit{AKDA} & \textit{KSVM} && \textit{KSDA} & \textit{GSDA}
     & \textit{AKSDA}\\
\midrule
  \textit{AwA}            & 1.14/1.1 & 2.34/\textbf{1.4} & 1.92/0.5 && 1/1 & 0.29/0.6 & 1.06/0.6 & \textbf{3.11}/1 & 0.10/0.3 && 0.01/1 & 0.41/0.6 & 1.9/0.8 \\
  \textit{ayahoo}       & 1.18/1 & 0.05/\textbf{1.2} & 0.53/0.7 && 1/1 & 0.96/0.9 & 2.06/0.9 & \textbf{2.56}/1 & 0.46/0.7 && 0.10/1 & 0.50/0.9 & 1.65/1 \\
  \textit{bing}            & 4.54/1.8 & 5.11/\textbf{2.5} & 2.65/0.9 && 1/1 & 0.52/0.4 & 9.74/0.5 & \textbf{21.8}/1 & 0.32/0.1 && 0.04/1 & 0.81/0.4 & 20.2/1  \\
  \textit{caltech101} & 2.11/1.3 & 3.4/\textbf{1.7} & 0.83/0.9 && 1/1 & 0.63/0.7 & 2.79/0.8 & \textbf{7.88}/1 & 0.32/0.3 && 0.02/1 & 1.04/0.7 & 6.97/1 \\
  \textit{caltech256} & 4.11/1.7  & 4.8/\textbf{2.3} & 3.26/1 && 1/1 & 0.53/0.5 & 7.16/0.4 & \textbf{21.8}/1 & 0.75/0.3 && 0.04/1 & 0.81/0.4 & 20/0.9  \\
  \textit{eth80}         & 0.15/1.3 & 3.17/\textbf{1.6} & 0.9/1 && 1/1 & 0.68/0.8 & 2.92/0.8 & \textbf{7.24}/1 & 0.87/1 && 0.02/1 & 1.05/0.8 & 5.93/1\\
  \textit{imagenet}   & 2.26/1.3 & 3.48/\textbf{1.7} & 0.81/0.7 && 1/1 & 0.52/0.7 & 4.07/0.7 & \textbf{9.53}/1 & 0.17/0.2 && 0.02/1 & 0.83/0.7 & 8.31/1 \\
  \textit{mscorid}      & 0.97/1 & 1.88/\textbf{1.2} & 0.26/0.8 && 1/1 & 0.49/0.9 & 1.12/0.9 & \textbf{1.62}/1 & 0.17/0.6 && 0.02/1 & 0.56/0.9 & 1.17/1\\
  \textit{office}         & 2.15/1.3 & 3.43/\textbf{1.6} & 0.73/0.9 && 1/1 & 0.63/0.8 & 3.09/0.8 & \textbf{7.63}/1 & 0.25/0.4 && 0.02/1 & 0.93/0.8 & 6.21/1 \\
  \textit{pascal07}    & 2.19/1 & 0.44/\textbf{1.2} & 2.36/0.6 &&  1/1 & 1/0.9 & 1.03/0.8 & \textbf{3.44}/1 &  0.15/0.01 &&  0.06/1 & 1.09/0.9 & 2.35/1 \\
  \textit{rgdb}           & 1.68/1.1 & 2.76/\textbf{1.4} &  0.47/0.8 && 1/1 & 0.58/0.8 & 1.89/0.9 & \textbf{4.73}/1 & 0.25/0.5 && 0.02/1 & 0.91/0.8 & 3.62/0.5 \\
\midrule
  \textit{Average}    & 2.04/1.3 & 2.81/\textbf{1.6} & 1.34/0.8 && 1/1 & 0.62/0.7 & 3.36/0.7 & \textbf{8.26}/1 & 0.35/0.4 && 0.03/1 & 0.81/0.7 & 7.11/0.9 \\
\bottomrule
\end{tabular}}
\end{center}
\label{tbl:crossDtStTrnTstTimes10Ex}
\end{table*}
\begin{table*}[!htb]
\caption{Training/testing time speedup on cross-dataset collection  -- 100Ex condition}
\begin{center}
{\footnotesize
\begin{tabular}{lccccccccccccc}
\toprule
     & \multicolumn{3}{c}{Linear methods} && \multicolumn{5}{c}{Kernel methods}
    && \multicolumn{3}{c}{Kernel subclass methods} \\
    \cmidrule{2-4} \cmidrule{6-10} \cmidrule{12-14}
     & \textit{PCA} & \textit{LDA} & \textit{LSVM} && \textit{KDA} & \textit{GDA} 
     & \textit{SRKDA} & \textit{AKDA} & \textit{KSVM} && \textit{KSDA} & \textit{GSDA}
     & \textit{AKSDA}\\
\midrule
  \textit{AwA}            & 5.17/2 & 40.8/\textbf{3.7} & 3.04/0.6 && 1/1 & 0.62/0.3 & 20.1/0.3 & \textbf{46.3}/1 & 1.27/0.3 && 0.07/1 & 1.02/0.3 & 43.1/1 \\
  \textit{ayahoo}       & 1.46/1.3 & 3.24/\textbf{1.7} & 1.46/1.3 && 1/1 & 0.6/0.7 & 3.4/0.8 & \textbf{7.78}/1 & 0.29/0.1 && 0.02/1 & 1.04/0.7 & 5.62/1 \\
  \textit{bing}            & 202/7 & \textbf{3756}/\textbf{15} & 68/1.2 && 1/1 & 1.3/0.1 & 151/0.1 & 258/1 & 10/0.3 && 0.32/0.3 & 2.1/0.1 & 252/1 \\
  \textit{caltech101} & 3.87/2.2 & 6.1/\textbf{3.2} & 5.2/1.2 && 1/1 & 0.75/0.4 & 15.4/0.4 & \textbf{33.2}/1 & 2.68/0.7 && 0.06/1 & 1.25/0.4 & 32.7/1 \\
  \textit{caltech256} & 60.8/4.4 & \textbf{517}/\textbf{7.9} & 26.5/1 && 1/1 & 0.64/0.1 & 50.5/0.1 & 101/1 & 14/0.7 && 0.18/0.8 & 1.04/0.1 & 99.5/1 \\
  \textit{eth80}         & 0.7/1.7 & 5.04/\textbf{2} & 2.28/1 && 1/1 & 1.06/0.6 & 7.3/0.6 & \textbf{16.2}/1 & 2.25/0.4 && 0.03/1 & 1.6/0.6 & 15/1 \\
  \textit{imagenet}   & 45.1/3.5 & \textbf{368}/\textbf{7.4} & 16.3/1 && 1/1 & 0.7/0.1 & 49/0.1 & 93.6/1 & 2.02/0.3 && 0.67/0.97 & 1.07/0.1 & 91.9/1 \\
  \textit{mscorid}      & 2.79/1.5 & 4.19/\textbf{2} & 0.99/0.7 && 1/1 & 0.5/0.6 & 5.5/0.6 & \textbf{12.1}/1 & 0.46/0.4 &&  $10^{-4}$/1 & 0.76/0.6 & 10.9/1 \\
  \textit{office}         & 1.88/1.8 & 5.95/2.4 & 2.57/1 && 1/1 & 0.8/0.5 & 9.8/0.5 & \textbf{22.08}/1 & 0.78/0.4 && 0.04/1 & 1.21/0.5 & 20.1/1 \\
  \textit{pascal07}    & 2.1/0.5 & 0.01/0.5 & 2.51/\textbf{1.2} && 1/1 & 1/0.9 & 1.3/0.9 & \textbf{4.38}/1 & 0.005/0.8 && 0.08/1 & 1.72/1 & 4.17/1 \\
  \textit{rgdb}           & 6.84/2.5 & 47.58/\textbf{4} & 8.2/1.3 && 1/1 & 1.38/0.3 & 23.3/0.3 & \textbf{49.4}/1 & 3.27/0.61 && 0.08/1.02  & 2.14/0.3 & 47.8/1 \\
\midrule
  \textit{Average}           & 30/2.5 & \textbf{432}/\textbf{4.5} & 12/1 && 1/1 & 0.8/0.4 & 31/0.4 & 58/1 & 3.4/0.5 && 0.16/1 & 1.4/0.4 & 56/1  \\
\bottomrule
\end{tabular}}
\end{center}
\label{tbl:crossDtStTrnTstTimes100Ex}
\end{table*}

The evaluation results in terms of MAP and training/testing time speedup for the MED datasets and the cross-dataset collection are shown in
Tables \ref{tbl:trecMedMap}, \ref{tbl:crossDtStMap10Ex}, \ref{tbl:crossDtStMap100Ex},
\ref{tbl:trecMedTrnTstTimes}, \ref{tbl:crossDtStTrnTstTimes10Ex}, and \ref{tbl:crossDtStTrnTstTimes100Ex}.
Starting our analysis from the linear DR methods,
we see that LDA  outperforms PCA only in the two MED datasets and in the larger datasets of the cross-dataset collection under the 100Ex condition.
Moreover, we observe that the gain of LDA over PCA is increased with the size of the dataset.
For instance, this is the case for caltech256, bing, med--hbb and imagenet,
where an 8\%, 5\%, 4\% and 3\% MAP improvement is observed, respectively.
This is explained by the fact that in most experiments
the small sample size (SSS) problem is intense, i.e. the number of training observations is much less than the dimensionality of the input space.
Due to this problem, the within-class scatter matrix is severely ill-posed, and, thus,
LDA fails to effectively identify a discriminant subspace.
Moreover, in general we observe that the use of the linear DR step (LDA or PCA) has a negative effect in the 
performance of the LSVM.
This shows that in most experiments we have to deal with dense nonlinear problems,
and the exclusion of even few dimensions (as with PCA) causes a quite significant loss of information.
We should note, that an exception to this rule is the med10 dataset,
where discarding noise or non-discriminant dimensions in the very high-dimensional input space aids the classification performance.


Moving forward to the analysis of (class-based) nonlinear methods (KDA, GDA, SRKDA, AKDA and KSVM), in average we observe 
a small improvement of approximately 1\% MAP over the best linear method in the MED datasets and the cross-datasets under the 10Ex condition,
and a more significant improvement of more than 2\% MAP in the cross-datasets under the 100Ex condition.
That is, the nonlinear DR approaches achieve to mitigate dataset nonlinearities to a satisfactory degree.
We also observe that in contrast to the linear case, kernel DR approaches
have a positive effect on LSVM's MAP rate, outperforming both LSVM and KSVM.
This is because, in addition to dealing with dataset nonlinearities,
these methods alleviate the curse-of-dimensionality problem in LSVMs 
by discovering a lower dimensional subspace where classes are linearly separable \cite{Vapnik98}.
More specifically, lower-capacity LSVM classifiers can now be effectively applied in the discriminant subspace,
improving the overall generalization performance of the combined classifier.
Finally, we observe that AKDA provides the best retrieval performance among all class-based kernel methods.
For instance, in the cross-datasets under the 100Ex condition we observe
an average MAP rate improvement of more than 2\% over GDA, KDA and KSVM.
This justifies the very good numerical properties of AKDA as explained in Section \ref{SS:simRdctAkda}.
That is, in contrary to computing and operating on large scatter matrices,
AKDA consists of few elementary matrix operations and very stable decomposition methods.
Similar conclusions to the one presented above for the class-based nonlinear methods
can be drawn for their subclass-based counterparts, i.e., KSDA, GSDA and AKSDA.
Additionally, we observe that the subclass-based methods achieve a small but noticeable gain over the class-based ones.
We can also see that in average, the proposed AKSDA outperforms KSDA by 2\% MAP, and in overall,
achieves the best retrieval performance among all approaches.

The major advantage of the proposed methods is the training time improvement over their conventional
counterparts, KDA and KSDA.
To this end, we observe an impressive speedup of AKDA over KDA ranging from $1.62\times$
for the mscorid dataset in the 10Ex task, to $258\times$ for the bing dataset in the 100Ex task.
This improvement is even larger in the comparison between KSDA and AKSDA,
where a training time speedup ranging from $15.8\times$ (mscorid, 10Ex) to $788\times$ (bing, 100Ex) is attained.
For instance, the estimated training time on one CPU core for both the proposed methods in the bing dataset and 100Ex condition,
i.e. for learning 257 classes using 25698 training observations for each class, is approximately 8 hours,
while the respective times for KDA and KSDA are 91 and 284 days, respectively.
These times could be further suppressed using conventional parallel computing architectures,
as shown in \cite{Chartampilas15,Chartampilas16}, where impressive speedups are reported
using an 8-core machine with a low-end GPU.
Furthermore, we observe that the proposed methods are several times faster than LSVM, KSVM and linear PCA,
and approximately 2 times faster than SRKDA (which is the previous state-of-the-art) in all experiments.
In overall, they achieve the best training times in 20 out of 24 experiments,
and only outperformed (in training time) by LDA in the 3 larger datasets, namely,
bing, caltech256 and imagenet under the 100Ex condition. 

Concerning testing times, the linear DR methods (PCA and LDA) attain the best performance in most experiments.
Due to the kernel function evaluations executed at the testing stage as shown in (\ref{E:kernelRpr}), 
a somewhat higher testing times are observed for most kernel-based methods,
including KDA, KSDA, and the proposed AKDA and AKSDA.
GDA, SRKDA and GSDA yield the worst testing time performance, mainly due to the additional cost imposed
for centering the test observations in the feature space using (\ref{E:prjTstObsCntrd}).

\section{Conclusions} \label{S:conclusions}

In this paper, using a novel factorization and simultaneous reduction
framework, two efficient variants of KDA, namely AKDA and AKSDA,
are presented. Specifically, based on very stable decomposition algorithms
and dealing with relatively smaller size matrices,
the proposed approaches exhibit excellent numerical properties
and more than one order of magnitude
computational complexity reduction in comparison to KDA and KSDA.
Moreover, experimental results show that combined with
LSVM, the proposed methods achieve state-of-the-art
performance in terms of both training time
and classification accuracy.

The proposed acceleration framework is general and may be worth
investigating its applicability to other techniques, such as,
incremental discriminant analysis, Laplacian eigenmaps and support vector machines.
Possible future research directions include the investigation of the above possibilities 
and the application of the proposed framework in other application domains
such as  large-scale visualization and recursive learning.

\ifCLASSOPTIONcompsoc
  \section*{Acknowledgments}
\else
  \section*{Acknowledgment}
\fi

This work was supported by the EU's Horizon 2020 programme under grant agreement H2020-693092 MOVING.






\bibliographystyle{IEEEtran}

\begin{IEEEbiography}[]{Nikolaos Gkalelis}
\end{IEEEbiography}

\begin{IEEEbiography}[]{Vasileios Mezaris}
\end{IEEEbiography}

\end{document}